\documentclass[10pt,twocolumn,letterpaper]{article}

\usepackage{iccv}              % To produce the CAMERA-READY version
\usepackage[dvipsnames]{xcolor}
\usepackage[accsupp]{axessibility}  % Improves PDF readability for those with disabilities.

\definecolor{iccvblue}{rgb}{0.21,0.49,0.74}
\usepackage[pagebackref,breaklinks,colorlinks,allcolors=iccvblue]{hyperref}

\def\paperID{9709} % *** Enter the Paper ID here
\def\confName{ICCV}
\def\confYear{2025}

\title{Referring Expression Comprehension for Small Objects}
\usepackage{tcolorbox}
\usepackage{minted}
\usepackage{amsmath,amssymb}
\usepackage{algorithmic}
\usepackage{algorithm}
\usepackage{bm}
\usepackage{booktabs}
\usepackage{comment}
\usepackage{multirow}
\usepackage{framed}
\usepackage{wrapfig} % added
\def\red#1{{\textcolor{red}{#1}}}
\def\blue#1{{\textcolor{blue}{#1}}}
\usepackage{here}
\usepackage{url}
\makeatletter
\newcommand{\figcaption}[1]{\def\@captype{figure}\caption{#1}}
\newcommand{\tblcaption}[1]{\def\@captype{table}\caption{#1}}
\makeatother

\newcount\K
\def\bla#1{
\K=0 \loop\ifnum\K<#1
{\textcolor[gray]{0.9}{{\it bla bla bla bla bla bla bla bla bla bla bla bla bla bla bla}}}
\advance\K by1\repeat
}

\newcommand{\argmin}{\mathop{\rm argmin}\limits}

\usepackage{float}

\author{
Kanoko Goto\textsuperscript{1}\thanks{Equal contribution} \quad
Takumi Hirose\textsuperscript{1}\footnotemark[1] \quad
Mahiro Ukai\textsuperscript{1} \quad
Shuhei Kurita\textsuperscript{2,1} \quad
Nakamasa Inoue\textsuperscript{1}
\\[2ex]
\textsuperscript{1}Institute of Science Tokyo \quad
\textsuperscript{2}National Institute of Informatics
\\[2ex]
{\tt\small Project Page: \href{https://github.com/mmaiLab/sorec}{https://github.com/mmaiLab/sorec}}
}

\begin{document}
\maketitle
\begin{abstract}
Referring expression comprehension (REC) aims to localize the target object described by a natural language expression.
Recent advances in vision-language learning have led to significant performance improvements in REC tasks.
However, localizing extremely small objects remains a considerable challenge despite its importance in real-world applications such as autonomous driving.
To address this issue, we introduce a novel dataset and method for REC targeting small objects.
First, we present the small object REC (SOREC) dataset, which consists of 100,000 pairs of referring expressions and corresponding bounding boxes for small objects in driving scenarios.
Second, we propose the progressive-iterative zooming adapter (PIZA), an adapter module for parameter-efficient fine-tuning that enables models to progressively zoom in and localize small objects.
In a series of experiments, we apply PIZA to GroundingDINO and demonstrate a significant improvement in accuracy on the SOREC dataset.
Our dataset, codes and pre-trained models are publicly available on the project page.
\end{abstract}
    
\section{Introduction}
\label{sec:intro}
Object localization in images has been a long-term research topic in the field of computer vision.
Early studies introduced image datasets such as Pascal VOC~\cite{Everingham2010voc} and COCO~\cite{lin2014coco}, which involve bounding box annotations for predefined object categories, leading to the development of object detection models including CNN-based  models~\cite{girshick2014rcnn,girshick2015fastrcnn,Ren2015fasterrcnn,He2017maskrcnn} and Transformer-based models~\cite{Carion20detr, zhu2020deformabledetr, kamath2021mdetr, liu2022dabdetr, liu2023dqdetr, zhang2023codetr, zhang2023dino, lv2024rtdetr, Hou2024SalienceDETR}.
\begin{figure}[t]
\centering
\includegraphics[width=\linewidth]{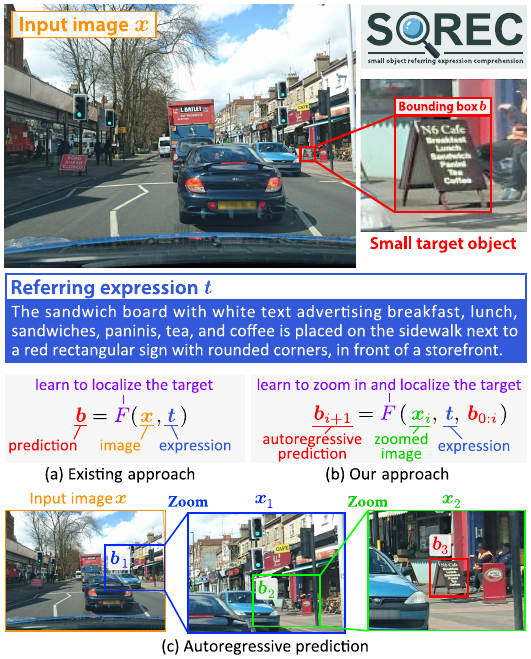}
\vspace{-20pt}
\caption{The SOREC dataset consists of pairs of referring expressions and bounding boxes for extremely small objects.
(a) Existing approach fine-tunes a model $F$ to localize the target.
(b) Our approach fine-tunes $F$ to progressively zoom in and localize the target in an autoregressive manner.
(c) Example of prediction in three zooming steps.}
\label{fig:teaser}
\vspace{-12pt}
\end{figure}
For more detailed and flexible object localization, referring expression comprehension (REC) aims to localize a specific object referred to by a natural language description.
REC uses queries like ``the red car parked in front of the coffee shop'' as input and requires locating this unique object in an input image.
RefCOCO, RefCOCO+ and RefCOCOg~\cite{yu2016refcoco, mao2016refcocog} are the most popular datasets for REC, providing referring expressions for images in the COCO dataset along with Flickr30K entities~\cite{plummer2015flickr30k}.
Over the past decade, deep learning architectures that bridge visual contents and natural language descriptions have been investigated. Examples include one-stage architectures~\cite{sadhu2019zero, liao2020real, luo2020multi, yang2019fast} and two-stage architectures based on CNN-LSTM~\cite{mao2016generation, yu2016refcoco, nagaraja2016modeling, hu2016natural, luo2017comprehension, liu2017referring, yu2017joint, zhang2018grounding} and attention mechanism~\cite{hu2017modeling, yu2018mattnet, zhuang2018parallel, deng2018visual, liu2019improving, luo2020multi, Song_2021_CVPR, Ding2024revisit}.

With advances in large-scale vision-language learning, recent models have become capable of precisely understanding object attributes and relations described in natural language~\cite{Li2022GLIP, zhang2022glipv2, wang2022ofa, yan2023uninext, wang2023one, Liu2024GroundingDINO, zhang2024mmgroundingdino}.
As a result, these models have achieved high accuracy in REC tasks, with accuracy rates of over 90\% on the RefCOCO test sets.
However, localizing small objects remains a significant challenge.
The lack of REC datasets targeting small objects has impeded further progress in this area, despite its critical role in real-world applications such as autonomous driving, where the ability to localize small objects is essential for ensuring safety and facilitating precise decision-making in complex environments.

To address this issue, the present study makes two significant contributions.
First, we introduce the small object REC (SOREC) dataset, a new dataset for REC targeting small objects in autonomous driving scenarios.
Second, we propose the progressive-iterative zooming adapter (PIZA), an adapter module for parameter-efficient fine-tuning that enables models to progressively zoom in and localize small objects.
Below we highlight each contribution.

\noindent \textbf{1) Dataset contribution.} We propose the SOREC dataset, which consists of 100,000 pairs of referring expressions and corresponding bounding boxes for extremely small objects in road, highway, rural-area, and off-road scenes.
As shown in Figure~\ref{fig:teaser},
the typical size of a bounding box is approximately 0.1\% of the input image size, making it challenging to localize these objects.
To the best of our knowledge, this is the first dataset for REC targeting small objects in autonomous driving scenarios.
The availability of this dataset promotes further research advancements in the area.

\noindent \textbf{2) Technical contribution.}
We propose PIZA, a lightweight learnable module for parameter-efficient fine-tuning.
Through fine-tuning with PIZA, the model learns to localize small objects in an autoregressive manner, where a zoomed image is fed into the model iteratively, as shown in Figure~\hyperref[fig:teaser]{\ref*{fig:teaser} (b-c)}.
This approach significantly improves accuracy, as demonstrated in Table~\ref{tab:main_results}.

\section{Related work}
\label{sec:related_work}

\subsection{Tasks and datasets}

\noindent \textbf{Referring expression comprehension.}
Considerable efforts have been dedicated to constructing datasets of referring expressions on images over the past decade. 
ReferItGame~\cite{kazemzadeh2014referitgame} was a pioneering large-scale dataset for REC, consisting of 130k expressions for 20k images collected from ImageCLEF and SAIAPR.
RefCOCO, RefCOCO+~\cite{yu2016refcoco} and RefCOCOg~\cite{mao2016refcocog} provided expressions for images in COCO~\cite{lin2014coco}, and are the most popular benchmarking datasets.
CLEVR-Ref~\cite{liu2019clevrref} is a diagnostic dataset focusing on compositional language understanding using synthetic images.
Refer360~\cite{xie2021refer360} is a dataset for referring expression recognition in 360-degree images.
REVERIE~\cite{qi2020reverie} offers a dataset for remote embodied visual referring expressions in real indoor environments.
RefEgo~\cite{Kurita2023refego} focuses on egocentric REC in first-person videos.

\noindent \textbf{Small object detection.}
The importance of small object detection has been recognized in various object detection scenarios.
Examples of datasets involving small objects include
WiderFace~\cite{yang2016wider} for face detection,
TinyPerson~\cite{yu2020tinyperson} for person detection,
TT100K~\cite{zhu2016tt100k} for traffic sign detection,
VisDrone~\cite{zhu2021VisDrone} for drone-based detection, and DOTA~\cite{ding2022dota} for remote sensing detection.
SODA~\cite{cheng2023sodad} is the latest large-scale dataset for small object detection in automatic driving scenarios.
Compared with normal-sized object detection datasets such as Pascal VOC~\cite{Everingham2010voc} and COCO~\cite{lin2014coco}, creating datasets for small object detection is generally more expensive as annotating small objects is more challenging.

\noindent \textbf{Other related tasks.}
Visual grounding (VG) also aims to localize objects given natural language descriptions.
In VG, each image may contain multiple target objects, typically described with shorter phrases than those used in REC.
Example VG datasets include SK-VG~\cite{chen2023skvg} and GigaGrounding~\cite{Ma2024gigagrounding}.
Open-vocabulary object detection aims to detect objects that are not seen in the training dataset.
For training open-vocabulary detection models, large object detection datasets are recently used such as O365~\cite{Shao2019o365}, GoldG~\cite{kamath2021mdetr}, GRIT~\cite{peng2023kosmos} and V3Det~\cite{wang2023v3det}.

\begin{figure*}[t]
\centering
\includegraphics[width=\linewidth]{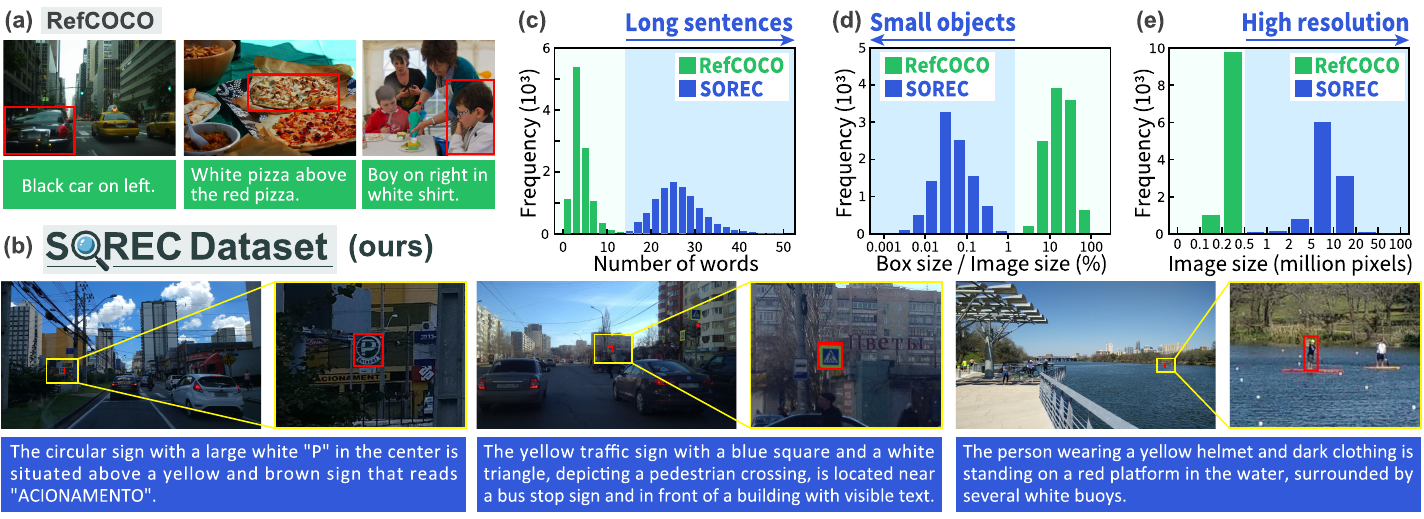}
\vspace{-18pt}
\caption{
Dataset comparison.
(a) RefCOCO is a representative REC dataset consisting of expressions and bounding boxes for normal-sized objects.
(d) SOREC is our dataset, consisting of relatively longer expressions compared to RefCOCO, to identify small objects.
(c-e) Comparison of word count, image size, and relative bounding box size distributions on test sets.
}
\vspace{-12pt}
\label{fig:sorec}
\end{figure*}

\subsection{Models}

REC models have evolved significantly over the years, transitioning from traditional CNN-LSTM architectures to attention and transformer-based architectures.
One-stage REC models~\cite{liao2020real, luo2020multi, yang2019fast, sadhu2019zero}
integrated object detection and language grounding into a unified architecture, allowing for end-to-end training.
Two-stage approaches~\cite{mao2016generation, yu2016refcoco, nagaraja2016modeling, hu2016natural, luo2017comprehension, liu2017referring, yu2017joint, zhang2018grounding} utilized region proposals generated by object detectors and applied LSTM to encode the referring expressions.
Attention mechanisms were later incorporated to improve the alignment between image regions and referring expressions~\cite{hu2017modeling, yu2018mattnet, zhuang2018parallel, deng2018visual, liu2019improving, luo2020multi, Song_2021_CVPR, Ding2024revisit}. 

To cover multiple vision tasks recent studies have demonstrated the effectiveness of large-scale vision-language pre-training~\cite{li2021glip, wang2022ofa, zhang2022glipv2, yan2023uninext, Liu2024GroundingDINO, zhang2024mmgroundingdino,wang2023one,dai2024simvg}.
For example, open-set detection models such as GLIPv2~\cite{zhang2022glipv2} and Grounding DINO~\cite{Liu2024GroundingDINO} can handle both object detection and REC.
We chose GroundingDINO as the baseline model because it is pre-trained on a union of datasets, including those for relatively small object detection and REC.

\subsection{Parameter efficient fine-tuning}

\noindent \textbf{Prompt-based fine-tuning.}
Inspired by prompt tuning methods for natural language processing tasks~\cite{shin2020autoprompt,jiang2020how,liu2022ptuning}, prompt-based fine-tuning methods have been proposed for computer vision tasks.
Context optimization (CoOp)~\cite{zhou2022coop} and visual prompt tuning (VPT)~\cite{jia2022vpt} are two representative methods.
CoOp incorporated learnable embeddings into the text encoder of CLIP~\cite{alec21CLIP}.
CoCoOp~\cite{zhou2022cocoop} introduced prompts conditioned by image features.
VPT incorporated learnable embeddings into the vision transformers~\cite{dosovitskiy2020vit}.
Further extension includes distribution learning, multi-modal learning and various techniques to leverage pre-trained knowledge~\cite{lu2022prompt, derakhshani2023bayesianprompt, Yao23KgCoOp, Khattak23Maple, Khattak23srp, Zhu23prograd,Cho23DAPT,Zhang24dept,yao24tcp}.

\noindent \textbf{LoRA-based fine-tuning.}
LoRA~\cite{hu2022lora} introduces low-rank adaptations to the weight matrices of a pre-trained model. LoRA reduces the number of trainable parameters by decomposing the weight updates into low-rank matrices.
For further improving parameter efficiency, quantization techniques are also introduced~\cite{dettmers2023qlora, xu2024qalora}.

\noindent \textbf{Adapter-based fine-tuning.}
Adapters are lightweight learnable modules incorporated into a frozen pretrained model.
The first adapter architecture~\cite{neil19nlpadapter} was proposed for Transformers, which inserts adapters into the attention module and the feed forward network module in each encoder layer.
There have been substantial efforts in architectural adapter design for various neural networks for computer vision tasks~\cite{chen2022vitadapter,Chen2022AdaptFormer,Jie2022FFT,Hao2023CMA,He2023SensitivityAware,Steitz2024adapterp}.
Adapter+~\cite{Steitz2024adapterp} is a well-designed adapter architecture for vision transformers, which we employ in our experiments.

\def\piza{\hspace{-1pt}\text{\scalebox{0.56}{\raisebox{0.44ex}{$\bigcirc$}}} \hspace{-4.3pt}{*}}
\def\pizas{\hspace{-1pt}\text{\scalebox{0.56}{\raisebox{0.44ex}{$\bigcirc$}}} \hspace{-4pt}{*}}
\def\pizal{\hspace{-1pt}\text{\scalebox{0.56}{\raisebox{0.44ex}{$\bigcirc$}}} \hspace{-5.2pt}{*}}

\begin{figure*}[t]
  \centering
  \begin{minipage}{0.6\linewidth}
    \centering
    \includegraphics[width=\linewidth]{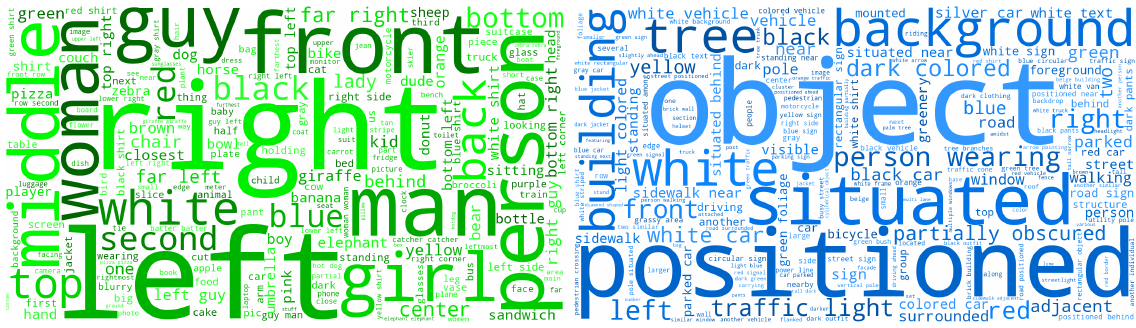}
    \vspace{-16pt}
    \caption{Word clouds for RefCOCO (left) and SOREC (right).}
    \label{fig:words}
    \vspace{-8pt}
  \end{minipage}
  \begin{minipage}{0.38\linewidth}
    \centering
    \vspace{12pt}
    \begin{tabular}{l|cc}
      \toprule
      Split & Images & Expressions \\
      \midrule
      Train-S & 1,446 & 10,000\\
      Train-L & 13,494 & 61,369\\
      Validation & 2,382 & 10,712 \\
      Test-A & 4,107 & 10,815\\
      Test-B & 5,153 & 17,104\\
      \bottomrule
    \end{tabular}
      \vspace{-6pt}
    \captionof{table}{Dataset split}
    \label{tab:split}
  \end{minipage}
  \vspace{-16pt}
\end{figure*}
\section{SOREC Dataset}

The SOREC dataset consists of 100,000 pairs of referring expressions and corresponding bounding boxes for small objects in road, highway, rural, and off-road images.
As shown in Figure~\hyperref[fig:sorec]{\ref*{fig:sorec} (b)}, referring expressions describe both the characteristics of the target object and its spatial relationships with surrounding objects, in order to locate target objects. Each bounding box typically occupies approximately 0.05\% of the entire image area.
Compared to existing datasets such as RefCOCO~\cite{yu2016refcoco} in Figure~\hyperref[fig:sorec]{\ref*{fig:sorec} (a)}, SOREC presents a particularly challenging task due to the extremely small bounding boxes. This challenge is critical for advancing real-world applications including autonomous driving and surveillance, where detecting small objects is~essential.

\subsection{Dataset Construction}

The SOREC dataset is semi-automatically created in the following five steps.

\noindent \textbf{1) Source selection.} We selected the SODA-D dataset~\cite{cheng2023sodad} as our source dataset. It consists of 24,828 high-quality images for small object detection, collected primarily from the Mapillary Vistas dataset~\cite{Neuhold2017mvd}. The average resolution of these images is $3407 \times 2470$ pixels.

\noindent \textbf{2) Segmentation.} To extract small object regions, we applied Semantic-SAM~\cite{li2024semanticsam} to image patches of size $800 \times 800$ pixels in a sliding window manner, with the granularity prompt level set to 3. Bounding boxes of object region were also computed. We excluded object regions whose bounding boxes occupied more than 2\% of the image.

\noindent \textbf{3) Filtering.} Through manual inspection of the extracted object regions, we found many instances of trees and windows that are not suitable for REC. We filtered them out by computing CLIP scores of each bound box region using a prompt of ``tree, forest, window.''
Subsequently, we sorted the object regions based on the score $S \exp(-|a - 1|) p$, where $S$ is the bounding box size, $a$ is the aspect ratio and $p$ is the stability score obtained from Semantic-SAM.
From the top 200,000 results, we excluded any remaining meaningless objects and selected the top 150,000 results through crowdsourcing.

\noindent \textbf{4) Referring expression generation.} For each object region, we cropped an image centered on its bounding box, with a random height between 2.5 and 3.5 times the height of the bounding box, and a width between 1.5 and 3 times its height.
We drew a red bounding box on the image with a line thickness of 2 pixels.
We input these images into GPT-4o to generate initial referring expressions using the prompt 
``Write a sentence that refers to the object in the red frame by mentioning its details, colors, relative relationship to surrounding objects.''
We excluded cases where the generated descriptions did not include references to surrounding objects by counting nouns.

\noindent \textbf{5) Quality control.} Finally, using crowdsourcing again, we excluded object regions that could not be uniquely identified by the corresponding expression.
This step was a binary decision on whether the object could be identified by the expression, allowing for minor errors in descriptions of surrounding objects, and chose 100,000 high-quality pairs of bounding boxes and expressions.
For test sets, we further asked annotators to revise expressions if they involve errors.
As a result, 18.45\% of sentences were found to contain minor errors related to color, spatial relations, and similar attributes.

\subsection{Dataset statistics}
\label{sec:statistics}
\noindent \textbf{Description length.}
Figure~\hyperref[fig:sorec]{\ref*{fig:sorec} (c)} shows the distribution of the number of words per expression. As shown, the average number of words is 25.5, which is approximately seven times longer than the 3.52 words in RefCOCO.
This is due to the need for more detailed information to identify small objects, which is a key characteristic of this dataset.
Figure~\ref{fig:words} compares word distributions by word clouds. As shown, SOREC dataset contains words related to positioning, such as 'positioned' and 'situated'.

\noindent \textbf{Bounding box size.}
Figure~\hyperref[fig:sorec]{\ref*{fig:sorec} (d)} shows the distribution of bounding box sizes relative to the image sizes. As indicated, all the target bounding boxes occupy less than 1\% of the image area, presenting a challenging REC task. Since pre-training is often performed on datasets that feature normal sized objects, fine-tuning is necessary to bridge this gap for localizing small target object.

\noindent \textbf{Image size.}
Figure~\hyperref[fig:sorec]{\ref*{fig:sorec} (e)} shows the distribution of image sizes. As shown, the SOREC dataset consists of high-resolution images that are sufficient for performing REC targeting small objects.

\noindent \textbf{Data split.}
We created training, validation, test splits as summarized in Table~\ref{tab:split}.
The train-L set is the full training set, and the train-S set is a small subset consisting of 10,000 expressions.
The validation set consists of 10,712 expressions, with no overlap in images with the training sets.
The test-A and -B sets contain expressions for traffic objects and the other objects.
\def\MM{\scalebox{0.8}{M}}

\section{Method}

This section describes PIZA, a lightweight adapter module for parameter-efficient fine-tuning that enables models to localize target small objects by progressively and iteratively zooming into them.

\subsection{Preliminary}

\noindent \textbf{Problem settings.}
Let $\bm{x} \in \mathbb{R}^{W \times H \times C}$ be an input image, where $W$ is the width, $H$ is the height, and $C$ is the number of channels.
We denote by $\bm{b} = (x_{0}, y_{0}, x_{1}, y_{1}) \in \mathbb{R}^{4}$ a bounding box, where $(x_{0}, y_{0})$ is the coordinate of the top-left corner, and $(x_{1}, y_{1})$ is the coordinate of the bottom-right corner of the box.
Given a natural language expression $\bm{t}$, the goal of REC is to localize the object corresponding to $\bm{t}$ in the image.
As such, the model $F$ takes as input $(\bm{x}, \bm{t})$ and learns to predict a bounding box as $\hat{\bm{b}} = F(\bm{x}, \bm{t})$, so that the predicted bounding box $\hat{\bm{b}}$ matches the ground truth bounding box $\bm{b}^{*}$.
This work focuses on the setting where the size of $\bm{b}^{*}$ is significantly smaller than the image size, \textit{i.e.}, we assume that $|\bm{b}^{*}| \ll WH$, where $|\bm{b}| = (x_{1} - x_{0})(y_{1} - y_{0})$ indicates the size of $\bm{b}$.

\noindent \textbf{Vision-language pre-training.} We assume that a pre-trained model $F$ is given and explore parameter-efficient fine-tuning methods by which the model quickly adapts to localize small objects.

\noindent \textbf{Difficulty.}
The main difficulty lies in the gap between pre-training and fine-tuning regarding the bounding box sizes and expression lengths.
Recent object localization models, such as GroundingDINO~\cite{Liu2024GroundingDINO}, are capable of both REC and open-set object detection because
they are pre-trained on a large union set of REC and object detection datasets.
However, a challenge arises when dealing with a combination of long sentences and extremely small objects. We aim to address this challenge in parameter-efficient fine-tuning.

\begin{figure}[t]
\centering
\includegraphics[width=\linewidth]{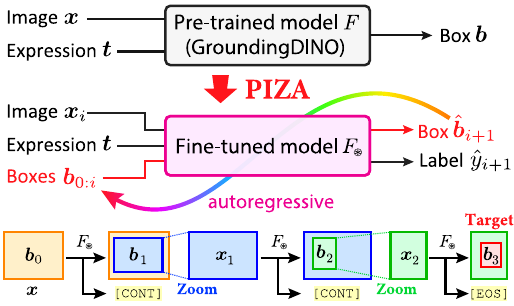}
\caption{\textbf{Fine-tuning with PIZA.} Given a pre-trained model $F$, PIZA produces a model $F_{\pizas}$ that zooms in to localize small objects in an autoregressive manner through fine-tuning.
In the inference phase, bounding boxes $\bm{b}_{0}, \bm{b}_{1}, \cdots, \bm{b}_{T}$ indicating zooming steps are predicted to localize the target at the end.
}
\label{fig:search_process}
\vspace{-12pt}
\end{figure}

\subsection{Progressive-integrative zooming adapter}

\noindent \textbf{Overview.}
Zooming in to localize a small target object can be understood as a search problem over an image. We model a search process $P$ as a sequence of bounding boxes:
\vspace{-5pt}
\begin{align}
P = (\bm{b}_{0}, \bm{b}_{1}, \cdots, \bm{b}_{T}),
\end{align}
where $\bm{b}_{0} = (0, 0, W, H)$ indicates the bounding box covering the entire image,
$\bm{b}_{T}$ is the final small bounding box for localizing the target, $T$ is the number of zooming steps, and $\bm{b}_{i}$ covers $\bm{b}_{j}$ if $i < j$, as shown in Figure~\hyperref[fig:teaser]{\ref*{fig:teaser} (c)}.

Through fine-tuning, models learn to predict search processes so that the final bounding box matches the ground truth, {\it i.e.}, $\bm{b}_{T} \simeq \bm{b}^{*}$.
To this end, PIZA extends the pre-trained model $F$ to a model $F_{\piza}$ that predicts a search process in an autoregressive manner as
\begin{align}
\label{eq:piza_ar}
\hat{\bm{b}}_{i+1} = F_{\piza}(\bm{x}_{i}, \bm{t}, \bm{b}_{0:i}),
\end{align}
where
$\bm{x}_{i}$ is the cropped image region corresponding to $\bm{b}_{i}$,
$\bm{t}$ is an input expression,
$\bm{b}_{0:i} = (\bm{b}_{0}, \cdots, \bm{b}_{i})$ is a subsequence of bounding boxes and $\pizal$ indicates that PIZA is applied.
A visualization of this procedure is shown in Figure~\ref{fig:search_process}.
Since $F$ is a function that accepts two inputs, $\bm{x}$ and $\bm{t}$, $F_{\piza}$ is built up by incorporating a module that can take $\bm{b}_{0:i}$ as an additional input into $F$. Below we describe the module architecture.

\begin{figure*}[t]
\centering
\hspace{-10pt}
\begin{minipage}{0.174\linewidth}
\includegraphics[width=\linewidth]{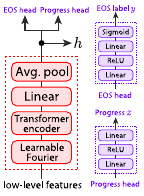}
\vspace{-12pt}
\caption{\hspace{-2pt}PIZA module.}
\label{fig:zoom_embedding}
\end{minipage}
\hspace{0pt}
\begin{minipage}{0.78\linewidth}
\includegraphics[width=\linewidth]{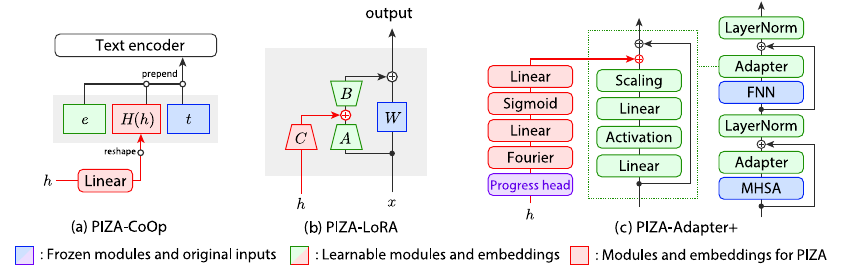}
\vspace{-17pt}
\caption{Parameter efficient fine-tuning with PIZA.}
\label{fig:variants}
\end{minipage}
\vspace{-12pt}
\end{figure*}

\noindent \textbf{PIZA module.}
Inspired by time-step embeddings in diffusion models~\cite{ho2020denoising, rombach2022sd} that represent stages of the diffusion process,
the PIZA module learns zooming-step embeddings that represent progress of the search process.
Specifically, the zooming-step embedding $\bm{h} \in \mathbb{R}^{d}$ is extracted from the input sequence of bounding boxes $\bm{b}_{0:i} \in \mathbb{R}^{4 \times (i+1)}$ in two steps.
First, a sequence of low-level features
$\bm{l}_{0:i} = (\bm{l}_{0}, \bm{l}_{1}, \cdots, \bm{l}_{i})$ is extracted.
Each feature $\bm{l}_{j}$ is a 6 dimensional vector, and its elements are listed in Table~\ref{tab:llf}.
Second, $\bm{h}$ is extracted by feeding $\bm{l}_{0:i}$ into a small learnable module. 
Figure~\ref{fig:zoom_embedding} shows the architecture consisting of a sequence of learnable Fourier embeddings~\cite{li2021llf}, a transformer encoder and an average pooling layer.
The embeddings are trained with two heads: an EOS head and a progress head.
The EOS head predicts a binary label $\hat{y}_{i+1} \in \{\texttt{[CONT]}, \texttt{[EOS]}\}$ indicating either ``continue to search (CONT)'' or ``end of search (EOS)''.
The progress head predicts progress of search $\hat{z}_{i+1} \in [0,1]$ expressed as a real value, where $0.0$ indicates the start and $1.0$ indicates the end of the search process.
The binary cross-entropy loss and mean squared error loss are applied to these heads, respectively, on the extended training dataset described in Section~\ref{sec:training}.
In the inference phase, search is stopped when the EOS label is predicted.
The number of parameters of this module is 0.27M and the feature dimension is set to 16.
The detailed architectural hyperparameters are provided with our code.

\begin{table}[t]
\centering
\small
\setlength{\tabcolsep}{10pt}
\begin{tabular}{l|l}
\toprule
Feature & Definition\\
\midrule
Normalized size & $s_{j} = |\bm{b}_{j}|/|\bm{b}_{0}|$ \\
Relative size & $r_{0}=1,~r_{j} = |\bm{b}_{j}|/|\bm{b}_{j-1}|$\\
Normalized width & $w_{j} = (x^{(j)}_{1} - x^{(j)}_{0})/W$\\
Normalized height & $h_{j} = (h^{(j)}_{1} - h^{(j)}_{0})/H$\\
Center position (x-axis) & $\bar{x}_{j} = (x^{(j)}_{0} + x^{(j)}_{1})/(2W)$\\
Center position (y-axis) & $\bar{y}_{j} = (y^{(j)}_{0} + y^{(j)}_{1})/(2H)$\\
\bottomrule
\end{tabular}
\vspace{-8pt}
\caption{Low-level features extracted from the sequence of bounding boxes $\bm{b}_{j} = (x^{(j)}_{0}, x^{(j)}_{1}, y^{(j)}_{0}, y^{(j)}_{1})$. $W$ and $H$ denotes the width and height of the input image.}
\label{tab:llf}
\vspace{-12pt}
\end{table}

\subsection{Parameter-efficient fine-tuning with PIZA}
\label{sec:piza_variants}

To perform fine-tuning with the PIZA module, we incorporate the embeddings $\bm{h}$ into parameter-efficient fine-tuning methods.
Here, we propose prompt-based, LoRA-based and adapter-based fine-tuning with PIZA. Their architectures are shown in Figure~\ref{fig:variants}.

\noindent
\textbf{PIZA-CoOp.}
CoOp~\cite{zhou2022coop} is a prompt-based fine-tuning method, which prepends learnable embeddings to the input text prompts as $G(\bm{x}, \bm{t}) = F(\bm{x}, [\bm{e}, \bm{t}])$, where $\bm{e} = (\bm{e}_{0}, \bm{e}_{1}, \cdots, \bm{e}_{L})$ is a sequence of learnable embeddings. PIZA-CoOp inserts $\bm{h}$ as $G_{\piza}(\bm{x}, \bm{t}) = F(\bm{x}, [\bm{e}, H(\bm{h}), \bm{t}])$, where $H$ is a learnable linear layer as shown in Figure~\hyperref[fig:variants]{\ref*{fig:variants}~(a)}.

\noindent \textbf{PIZA-LoRA.}
LoRA~\cite{hu2022lora} is a low-rank adaptation method that injects trainable low-rank matrices into linear projections as $W\bm{x}+BA\bm{x}$, where $\bm{x}$ is an input, $W$ is a frozen weight matrix and $A$, $B$ are learnable low-rank matrices.
PIZA-LoRA integrates the embeddings $\bm{h}$ into the bottleneck of LoRA as shown in Figure~\hyperref[fig:variants]{\ref*{fig:variants}~(b)}, resulting in $W\bm{x}+BA\bm{x} + BC\bm{h}$, where $C$ is a newly added learnable matrix. We set the rank to $16$.

\noindent \textbf{PIZA-Adapter+.}
Adapter+ \cite{Steitz2024adapterp} utilizes the post-adapter architecture, channel-wise scaling and Houlsby initialization~\cite{neil19nlpadapter}. PIZA-Adapter+ adds the embeddings $\bm{h}$ to the output of the channel-wise scaling layer as shown in Figure~\hyperref[fig:variants]{\ref*{fig:variants}~(c)}. We set the bottleneck dimension to 256.

\subsection{Training}
\label{sec:training}
Given a training dataset $\mathcal{D}$ consisting of images, expressions and ground truth bounding boxes, we construct an extended dataset $\mathcal{E}$ that involves ground truth search processes for training with PIZA.

\noindent \textbf{Overview.}
Each ground truth search process $P^{*}$ represents zooming steps to localize the target object.
Specifically, it is given by $P^{*} = (\bm{b}_{0}^{*}, \bm{b}_{1}^{*}, \cdots, \bm{b}_{T^{*}}^{*})$, where $\bm{b}_{0}^{*} = (0, 0, W, H)$ indicates the entire image and $\bm{b}_{T^{*}}^{*} = \bm{b}^{*}$ is the ground truth bounding box.
To facilitate efficient fine-tuning,
we generate $P^{*}$ so that the distribution of inverse zoom factors $(z^{*}_{j})^{-1} = |\bm{b}^{*}_{j}|/|\bm{b}^{*}_{j-1}|$ match to the distribution of bounding box area ratios $p(r)$ in pre-training.

\noindent \textbf{Distribution $\bm{p(r)}$.}
Along with a pre-trained model, the distribution $p(r)$ is pre-estimated by applying kernel density estimation to the pre-training dataset (we use the union of O365 and GoldG in our experiments).
We randomly sample 100,000 bounding boxes to compute area ratios $r = |\bm{b}^{*}|/(WH)$ and apply a Gaussian kernel. 

\noindent \textbf{Width and height for $\bm{b_{j}^{*}}$ .}
The width and height for each bounding box in $P^{*}$ are randomly sampled in three steps.
First, ratios $r_{k}$ are randomly sampled from $p(r)$ for $k = 1, 2, \cdots, T_{\text{max}}$, where $T_{\text{max}}$ is a sufficiently large constant.
Second, the number of zooming steps $T^{*}$ is determined such that the cumulative product of $r_{k}$ is closest the area ratio $r^{*}$ of the ground truth bounding box:
\begin{align}
T^{*}
=
\argmin_{T}
\left(\frac{1}{r^{*}} \prod_{k=1}^{T} r_{k} - 1 \right),\ 
r^{*} = \frac{|\bm{b}^{*}|}{WH},
\end{align}
where $\bm{b}^{*}$ is a bounding box sampled from $\mathcal{D}$, and $W, H$ are the width and height of the corresponding image.
Then, the zoom factors $z_{j}^{*}$ and the size of bounding boxes $S_{j}^{*}$ are determined as follow:
\begin{align}
z_{j}^{*}
\hspace{-3pt}=\hspace{-3pt}
\left(
\frac{1}{r^{*}}\prod_{k=1}^{T^{*}} r_{k}^{\omega_{k}^{-1}}\right)^{\hspace{-3pt}\frac{1}{T^{*}}}
\hspace{-9pt}
r_{j}^{-1},\,
S_{j}^{*}
\hspace{-3pt}=\hspace{-3pt}
\left( \prod_{k=1}^{T^{*}-j} z_{T^{*}-k} \right)  |\bm{b}^{*}|,
\end{align}
where $\omega_{k} = \lambda_{1} e^{-\lambda_{2} {k}}/\sum_{k'=1}^{T^{*}} \lambda_{1} e^{-\lambda_{2}{k'}}$ are weights derived from an exponential distribution.
This weighting ensures that $z_{j}^{*} \simeq r^{-1}_{j}$ with smaller $j$, encouraging more precise bounding box predictions at the initial zooming step.
Finally, the width and height of $\bm{b}_{j}^{*}$ are determined by
$
w^{*}_{j} = a_{j} \sqrt{S^{*}_{j}},\quad
h^{*}_{j} = a_{j}^{-1} \sqrt{S^{*}_{j}}
$
where $a_{j}$ is an interpreted aspect ratio
\begin{align}
a_{j}
=
\frac{W}{H}
-
\left(
\frac{W}{H} - 1
\right)
\frac{WH - S^{*}_{j}}{WH - S^{*}_{T^{*}-1}}.
\end{align}

\noindent \textbf{Centers.}
The center of each bounding box is aligned with the center coordinates of the target object. If the region extends beyond the image boundaries, the bounding box is minimally shifted to remain entirely within the image.

\noindent \textbf{Labels.}
Finally, for the generated search process $P^{*}$, the sequence of binary labels $\bm{y}^{*} = (y^{*}_{0}, y^{*}_{1}, \cdots, y^{*}_{T^{*}})$ and zooming step labels $z = (z^{*}_{0}, z^{*}_{1}, \cdots, z^{*}_{T^{*}},)$ are attached. Each label is given by
\begin{align}
y^{*}_{j} =
\begin{cases}
\texttt{[CONT]} & (0 \leq j < T^{*})\\
\texttt{[EOS]} & (j = T^{*})\\
\end{cases},\quad
z_{j}^{*} = \frac{j}{T^{*}}
\end{align}

\noindent \textbf{Loss function.}
The loss for fine-tuning is computed in three steps.
First, a mini-batch of quadruplets $(\bm{x}, \bm{t}, P^{*}, \bm{y}^{*})$ is drawn from $\mathcal{E}$.
Second, for each quadruplet, index $i \in \{1, 2, \cdots, T^{*}-1\}$ is randomly drawn to compute the forward process $\hat{\bm{b}}_{i+1}= F_{\piza}(\bm{x}, \bm{t}, \bm{b}_{0:i})$.
Finally, the loss depending on the pre-trained model is applied. In the experiments, we implement PIZA over the GroundingDINO model; thus the loss consists of the contrastive loss and the localization loss~\cite{Liu2024GroundingDINO}.

\def\metrics{\scalebox{0.87}{mAcc} & \scalebox{0.87}{Acc$_{50}$} & \scalebox{0.87}{Acc$_{75}$}}
\def\val#1#2#3#4{#4 & #2 & #3 &}
\def\testAB#1#2#3#4#5#6#7#8{#4 & #2 & #3 & #8 & #6 & #7 }
\def\testABx#1#2#3#4#5#6#7#8{\blue{#4} & \blue{#2} & \blue{#3} & \blue{#8} & \blue{#6} & \blue{#7}}
\def\subsets{\multicolumn{3}{c|}{Val} & \multicolumn{3}{c|}{Test-A} & \multicolumn{3}{c|}{Test-B} & \multicolumn{3}{c|}{Val} & \multicolumn{3}{c|}{Test-A} & \multicolumn{3}{c}{Test-B}}
\begin{table*}[t]
\small
\centering
\setlength{\tabcolsep}{1pt}
\begin{tabular}{l|c|ccc|ccc|ccc|ccc|ccc|ccc}
\toprule
\multirow{3}{*}{Method} & \multirow{3}{*}{\#Params} & \multicolumn{9}{c|}{Train-S} & \multicolumn{9}{c}{Train-L}\\
& & \subsets\\
& & \metrics & \metrics & \metrics & \metrics & \metrics & \metrics\\
\midrule
Zero-shot & 0 &
\val{4.0}{0.6}{0.0}{0.2}\testAB{6.4}{1.0}{0.1}{0.3}{1.8}{0.2}{0.0}{0.0} & \val{4.0}{0.6}{0.0}{0.2}\testAB{6.4}{1.0}{0.1}{0.3}{1.8}{0.2}{0.0}{0.0}\\
Full fine-tuning & 173.0M &
\val{58.6}{51.7}{30.2}{29.5}\testAB{62.5}{58.6}{38.8}{35.9}{50.6}{43.6}{21.9}{23.0} & \val{70.1}{\underline{63.7}}{39.2}{37.4}\testAB{72.8}{\underline{69.6}}{48.0}{43.8}{62.6}{\underline{55.6}}{29.8}{30.5}\\
\midrule
CoOp & 0.1M & \val{41.5}{36.1}{20.4}{20.2}\testAB{43.4}{40.1}{25.8}{24.2}{34.9}{29.6}{14.6}{15.5} & \val{48.8}{41.6}{22.0}{22.6}\testAB{51.1}{46.5}{28.7}{27.5}{42.1}{34.8}{15.9}{17.5}\\
PIZA-CoOp (Ours) & 0.9M &  \val{42.2}{39.1}{29.7}{26.3}\testAB{43.0}{41.2}{34.2}{29.4}{36.7}{33.8}{24.3}{21.9} & \val{47.5}{44.1}{33.5}{29.8}\testAB{48.6}{46.8}{38.7}{33.4}{40.8}{37.6}{26.9}{24.4}\\
\midrule
LoRA & 1.3M & \val{44.3}{38.5}{21.8}{21.6}\testAB{42.7}{43.1}{28.1}{26.2}{34.4}{32.5}{15.9}{17.0} & \val{51.0}{44.5}{25.3}{25.2}\testAB{54.1}{50.2}{33.0}{30.7}{44.3}{37.3}{18.8}{19.7} \\
PIZA-LoRA (Ours) & 1.5M &
\val{48.0}{44.7}{34.9}{30.9}\testAB{48.9}{46.6}{39.2}{33.8}{42.0}{38.7}{28.9}{25.8}
& \val{52.8}{49.9}{39.1}{34.5}\testAB{55.2}{54.0}{45.5}{39.3}{46.1}{43.4}{32.4}{29.0}\\
\midrule
Adapter+ & 3.3M & \val{55.4}{48.1}{24.8}{26.0} \testAB{59.4}{55.0}{33.3}{32.0}{48.3}{40.4}{17.9}{20.3} & \val{65.8}{59.5}{35.7}{34.6} \testAB{69.6}{65.9}{44.4}{40.7}{58.4}{51.3}{26.6}{27.6}\\
PIZA-Adapter+\hspace{1pt}(Ours) & 3.5M & \val{\underline{56.8}}{\underline{53.5}}{\underline{41.8}}{\underline{36.8}}\testAB{61.6}{\underline{59.6}}{\underline{50.1}}{\underline{43.1}}{\underline{49.5}}{\underline{45.9}}{\underline{34.1}}{\underline{30.4}} & \val{66.5}{60.6}{\underline{42.9}}{\underline{39.0}}\testAB{69.9}{66.2}{\underline{51.7}}{\underline{45.1}}{58.6}{52.2}{\underline{33.6}}{\underline{31.7}}\\
\bottomrule
\end{tabular}
\vspace{-8pt}
\caption{Parameter-efficient fine-tuning results. \#Params indicates the number of fine-tuned parameters. Best results are underlined.}
\vspace{-8pt}
\label{tab:main_results}
\end{table*}
\def\valmu#1#2#3#4{\scalebox{0.99}[1]{\underline{#4}/\underline{#2}/\underline{#3}}&}
\def\testABmu#1#2#3#4#5#6#7#8{\scalebox{0.92}[1]{\underline{#4}/\underline{#2}/\underline{#3}} & \scalebox{0.99}[1]{\underline{#8}/\underline{#6}/\underline{#7}}}
\def\valm#1#2#3#4{\scalebox{0.99}[1]{#4/#2/#3} &}
\def\testABm#1#2#3#4#5#6#7#8{\scalebox{0.92}[1]{#4/#2/#3} & \scalebox{0.99}[1]{#8/#6/#7}}

\def\result#1{\scalebox{0.93}[1]{#1}}
\def\resultz#1{\scalebox{0.98}[1]{#1}}

\section{Experiments}

\subsection{Experimental settings}

\noindent \textbf{Dataset and metrics.}
The SOREC dataset is used for training and evaluation.
We report the mean accuracy ($\text{mAcc}$) over IoU thresholds from 0.50 to 0.95 in increments of 0.05, as well as the accuracy at IoU of 0.50 ($\text{Acc}_{50}$) and 0.75 ($\text{Acc}_{75}$).

\noindent \textbf{Pre-trained model.}
We selected GroundingDINO~\cite{Liu2024GroundingDINO} using Swin-T~\cite{liu2021swin} as a baseline model, and used the improved version provided as MM-GroundingDINO in the mmdetection library~\cite{zhang2024mmgroundingdino}.
This model is pre-trained on the union of the following four datasets: O365~\cite{Shao2019o365}, GoldG~\cite{kamath2021mdetr}, GRIT~\cite{peng2023kosmos} and V3Det~\cite{wang2023v3det}.

\noindent \textbf{Baselines.}
We implemented three baselines for parameter-efficient fine-tuning: CoOp~\cite{zhou2022coop}, 
LoRA~\cite{hu2022lora} and Adapter+ \cite{Steitz2024adapterp}.
PIZA is applied to each method as described in Section~\ref{sec:piza_variants}.
Zero-shot baseline results are also reported.

\noindent \textbf{Implementation details.}
The AdamW optimizer is used for 5 epochs with a learning rate of $2 \times 10^{-4}$, which is decayed by a factor of 0.5 at epoch 3.
The hyperparameters for AdamW are set to their default values in PyTorch.
The batch size is set to 16. 
LoRA is applied to each self-attention and cross-attention module.
Adapter+ modules are inserted after each self-attention and feed-forward network module.
Further details are provided in Appendix.

\begin{figure*}[t]
\centering
\includegraphics[width=\linewidth]{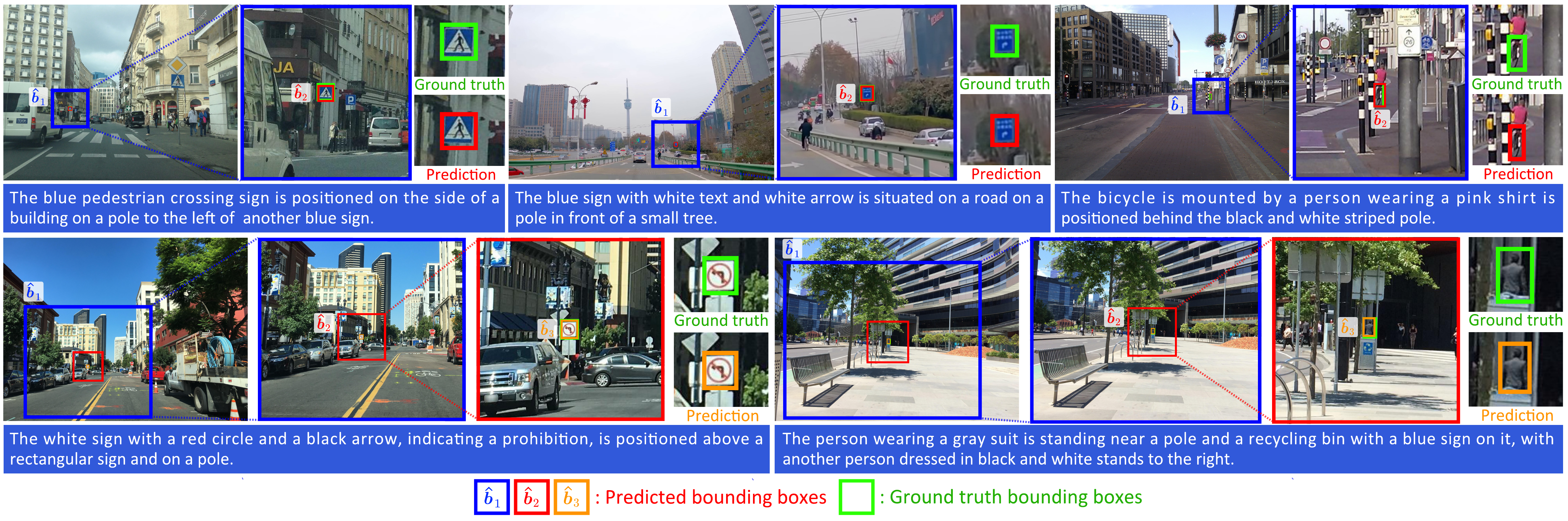}
\vspace{-17pt}
\caption{
Qualitative examples.
}
\vspace{-14pt}
\label{fig:qualitative}
\end{figure*}

\subsection{Experimental results}

\noindent \textbf{Main results.}
Table~\ref{tab:main_results} summarizes the parameter-efficient fine-tuning results.
As shown, PIZA significantly improved the performance for all methods.
PIZA-Adapter+ achieved the best performance in terms of mAcc, surpassing the full fine-tuning baseline while reducing the number of learnable parameters from 173.0M to 3.5M.
Prompt-tuning methods (CoOp and PIZA-CoOP) were more efficient but less effective than Adapter+ and LoRA methods.
This is likely due to the low performance of the zero-shot baseline, which suggests that prompt-tuning alone may struggle to bridge the gap between the vision-language pre-training task and the REC task for small objects. 
When comparing Test-A and Test-B, all models exhibited higher accuracy on Test-A, which consists of traffic objects. This result is understandable, as objects such as traffic lights and road signs are designed with colors and shapes that make them easily to detect.
When comparing training dataset sizes, the larger dataset (Train-L) consistently demonstrated higher performance, suggesting that further increasing the dataset size could be beneficial.

\begin{table}[t]
\begin{minipage}{0.485\textwidth}
\small
\setlength{\tabcolsep}{0.8pt}
\begin{tabular}{l|c|c|c|c}
\toprule
Method & \#Prm.  & Val & Test-A & Test-B \\
\midrule
PIZA-Adapter+ & \scalebox{0.97}[1]{3.5\MM} & \valmu{56.8}{53.5}{41.8}{36.8}\testABmu{61.6}{59.6}{50.1}{43.1}{49.5}{45.9}{34.1}{30.4}\\
\scalebox{0.94}[1]{w/o\hspace{1pt}emb.\hspace{1pt}insertion} & \scalebox{0.97}[1]{3.5\MM} & \valm{56.7}{53.2}{41.7}{36.7}\testABm{61.3}{59.2}{49.9}{42.8}{49.7}{45.8}{34.0}{30.3}\\
\scalebox{0.95}[1]{w/o\hspace{1pt}PIZA\hspace{1pt}module} & \scalebox{0.97}[1]{3.3\MM} & \valm{55.4}{48.1}{24.8}{26.0}\testABm{59.4}{55.0}{33.3}{32.0}{48.3}{40.4}{17.9}{20.3}\\
\midrule
$d = 256$ & \scalebox{0.97}[1]{3.5\MM} & \valm{56.8}{\underline{53.5}}{\underline{41.8}}{\underline{36.8}}\testABm{61.6}{\underline{59.6}}{\underline{50.1}}{\underline{43.1}}{\underline{49.5}}{\underline{45.9}}{34.1}{\underline{30.4}}\\
$d = 128$ & \scalebox{0.97}[1]{2.4\MM} & \valm{56.0}{52.8}{41.6}{36.4}\testABm{59.9}{57.9}{48.5}{41.8}{49.1}{45.5}{\underline{34.2}}{30.3}\\
$d = 64$ & \scalebox{0.97}[1]{1.9\MM} & \valm{56.2}{52.9}{\underline{41.8}}{36.6}\testABm{59.9}{58.1}{49.0}{42.2}{48.5}{45.1}{33.6}{29.9}\\
$d = 32$ & \scalebox{0.97}[1]{1.6\MM} & \valm{54.0}{51.0}{40.1}{35.1}\testABm{58.2}{56.3}{47.4}{40.8}{47.1}{43.7}{32.5}{29.0}\\
\bottomrule
\end{tabular}
\vspace{-8pt}
\caption{
Ablation and hyperparameter studies for PIZA-Adapter+.
``w/o embedding insertion'' omits the connection colored in red in Figure~\hyperref[fig:variants]{\ref*{fig:variants}\hspace{1pt}(c)}.\hspace{1pt}$d$\hspace{1.2pt}is the bottleneck\hspace{1.2pt}dimension\hspace{1.2pt}of the\hspace{1.2pt}adapter.\hspace{1pt}Each triplet\hspace{1pt}of values indicates \scalebox{1.0}{mAcc}/\scalebox{1.0}{Acc$_{50}$}/\scalebox{1.0}{Acc$_{75}$}. Train-S is used for training. 
}
\vspace{-8pt}
\label{tab:ablation_adapter}
\end{minipage}
\end{table}

\noindent \textbf{Ablation and hyperparameter studies.}
Tables~\ref{tab:ablation_adapter} and \ref{tab:ablation_lora} show the results of ablation and hyperparameter studies for PIZA-Adapter+, PIZA-LoRA, and PIZA-CoOp.
As shown, autoregressive prediction with the PIZA module is essential, and incorporating the zooming-step embedding further improved the performance.
For PIZA-Adapter+, performance improves as the bottleneck dimensions increase.

\noindent \textbf{Necessity of pre-training.}
Table~\ref{tab:scratch} compares training from scratch and full fine-tuning. Although the zero-shot baseline performance was low, the results confirmed that pre-training is necessary.

\noindent \textbf{Zooming steps.}
Table~\ref{tab:zooming} presents the tradeoff between the number of zooming steps and performance by comparing our best results, which resulted in 2.11 steps on average, with those obtained by enforcing the number of steps $T^{*}$ to 1, 2 and 3 when creating the extended training dataset.
As shown, our method performed the best among the tested configurations.
Some qualitative examples are shown in Figure~\ref{fig:qualitative}.
As shown, for the SOREC dataset, 2 or 3 zooming steps were sufficient in most cases.

\begin{table}[t]
\centering
\small
\setlength{\tabcolsep}{1pt}
\begin{tabular}{l|c|c|c|c}
\toprule
Method & \#Prm.  & Val & Test-A & Test-B \\
\midrule
PIZA-LoRA & \scalebox{0.97}[1]{1.5\MM} & \valmu{48.0}{44.7}{34.9}{30.9}\testABmu{48.9}{46.6}{39.2}{33.8}{42.0}{38.7}{28.9}{25.8}\\
\scalebox{0.95}[1]{w/o\hspace{1pt}emb.\hspace{1pt}insertion} & \scalebox{0.97}[1]{1.5\MM} & \valm{47.1}{43.9}{34.0}{30.2}\testABm{48.2}{46.4}{38.6}{33.5}{41.4}{38.1}{28.3}{25.3}\\
\scalebox{0.97}[1]{w/o\hspace{1pt}PIZA\hspace{1pt}module} & \scalebox{0.97}[1]{1.3\MM} & \valm{44.3}{38.5}{21.8}{21.6}\testABm{42.7}{43.1}{28.1}{26.2}{34.4}{32.5}{15.9}{17.0}\\
\midrule
PIZA-CoOp & \scalebox{0.97}[1]{0.9\MM} & \valmu{42.2}{39.1}{29.7}{26.3}\testABmu{43.0}{41.2}{34.2}{29.4}{36.7}{33.8}{24.3}{21.9} \\
\scalebox{0.95}[1]{w/o\hspace{1pt}emb.\hspace{1pt}insertion} & \scalebox{0.97}[1]{0.3\MM} & \valm{42.0}{38.7}{29.0}{26.1}\testABm{42.7}{40.9}{33.7}{29.3}{36.5}{33.2}{23.9}{21.6}\\
\scalebox{0.97}[1]{w/o\hspace{1pt}PIZA\hspace{1pt}module} & \scalebox{0.97}[1]{0.1\MM} & \valm{41.5}{36.1}{20.4}{20.2}\testABm{43.4}{40.1}{25.8}{24.2}{34.9}{29.6}{14.6}{15.5}\\
\bottomrule
\end{tabular}
\vspace{-8pt}
\caption{
Ablation study for PIZA-LoRA and PIZA-CoOp (Train-S). Each triplet of values indicates \scalebox{1.0}{mAcc}/\scalebox{1.0}{Acc$_{50}$}/\scalebox{1.0}{Acc$_{75}$}.
}
\vspace{-8pt}
\label{tab:ablation_lora}
\end{table}

\begin{table}[t]
\centering
\small
\setlength{\tabcolsep}{2.3pt}
\begin{tabular}{l|c|c|c|c}
\toprule
Method & \#Prm. & Val & Test-A & Test-B \\
\midrule
Scratch & 173\MM & \valm{0.00}{0.00}{0.00}{0.00}\testABm{0.00}{0.01}{0.00}{0.00}{0.00}{0.00}{0.00}{0.00}\\
Full fine-tuning & 173\MM & \valm{58.6}{51.7}{30.2}{29.5}\testABm{62.5}{58.6}{38.8}{35.9}{50.6}{43.6}{21.9}{23.0}\\
\bottomrule
\end{tabular}
\vspace{-8pt}
\caption{Comparison with training from scratch (Train-S).}
\vspace{-10pt}
\label{tab:scratch}
\end{table}

\begin{table}[t]
\vspace{2pt}
\begin{minipage}{0.485\textwidth}
\small
\setlength{\tabcolsep}{1.9pt}
\begin{tabular}{l|c|c|c|c}
\toprule
Method & Steps & Val & Test-A & Test-B\\
\midrule
PIZA-Adapter+ & 2.11 & \valmu{56.8}{53.5}{41.8}{36.8}\testABm{61.6}{\underline{59.6}}{\underline{50.1}}{\underline{43.1}}{\underline{49.5}}{\underline{45.9}}{\underline{34.1}}{\underline{30.4}}\\
\scalebox{0.95}[1]{w/ step enforcing} & 1.0 & \valm{54.6}{47.5}{25.5}{25.8}\testABm{58.5}{54.2}{33.3}{31.7}{47.0}{39.3}{18.1}{20.0}\\
\scalebox{0.95}[1]{w/ step enforcing} & 2.0 & \valm{56.9}{53.1}{40.8}{36.1}\testABm{60.2}{57.8}{48.6}{41.7}{49.5}{45.6}{32.9}{29.6}\\
\scalebox{0.95}[1]{w/ step enforcing} & 3.0 & \valm{54.5}{50.3}{39.0}{34.3}\testABm{58.0}{55.3}{46.3}{39.8}{46.2}{41.8}{30.1}{27.1} \\
\bottomrule
\end{tabular}
\vspace{-8pt}
\caption{Zooming step analysis (Train-S).}
\label{tab:zooming}
\vspace{-14pt}
\end{minipage}
\end{table}

\begin{figure}
\centering
\includegraphics[width=\linewidth]{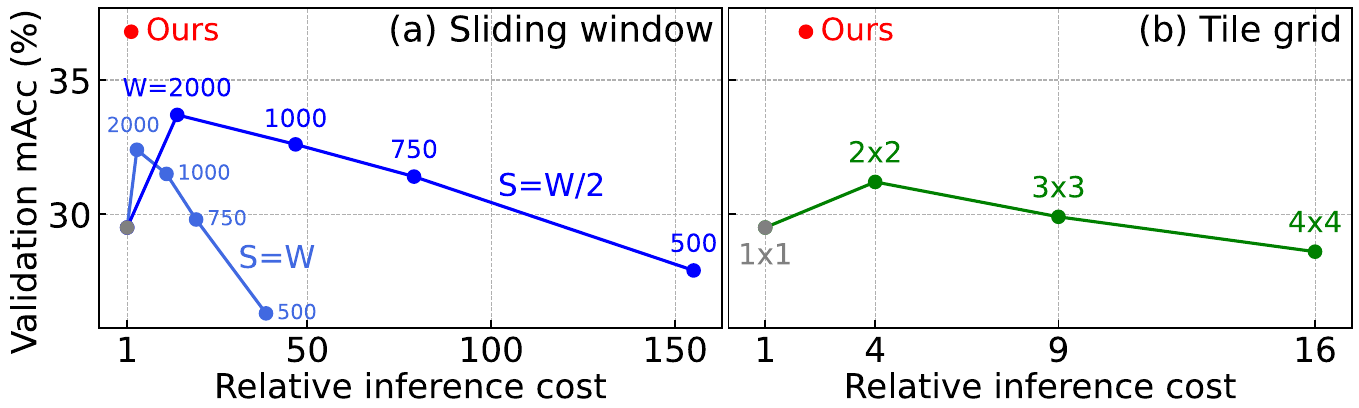}
\vspace{-18pt}
\renewcommand{\figurename}{Fig.}
\caption{\hspace{-2pt}Comparison with sliding window and tile-grid baselines.}
\vspace{-8pt}
\label{fig:sliding-window}
\end{figure}

\noindent \textbf{Comparison with greedy approaches.}
To validate the necessity of the multi-step inference, we compare PIZA with sliding window and grid-like separation approaches.
Figure~\hyperref[fig:sliding-window]{\ref*{fig:sliding-window}~(a)}
compares our method with the eight fully fine-tuned sliding window baselines using combinations of four window sizes $W\hspace{-1pt}\in\hspace{-1pt}\{500,750,1000,2000\}$ and two strides $S\hspace{-1pt}\in\hspace{-1pt}\{W, W/2\}$.
Apparently, the sliding window approach can improve the performance. However, all of these baselines indeed yield significantly lower performance and higher computational cost than our method.
Although smaller windows and smaller strides capture finer details, they exponentially increase computational cost and often lead to false positive detections.
Our method addresses these limitations, being 7.3$\times$ faster and using 49.4$\times$ fewer learnable parameters than the best sliding window baseline. 
Figure~\hyperref[fig:sliding-window]{\ref*{fig:sliding-window}~(b)}
shows comparison with the tile-grid baselines.
Our method outperformed them for the same reason discussed for the sliding window approach, suggesting that our method with average zooming step of 2.11 is reasonable compared to these greedy grid-like separation counterparts.
The typical window size used in the first step of our method was around $500\hspace{-1pt}\times\hspace{-1pt}500$, while 
that for the tile grid approach on a typical high-resolution image in our dataset was $3407\hspace{-1pt}\times\hspace{-1pt}2470$, resulting in a 36-fold increase in inference cost for the greedy tile grid approach.

\section{Conclusion}
We introduced the SOREC dataset, a new dataset for referring expression comprehension targeting small objects.
Furthermore, we proposed PIZA tuning, a novel parameter-efficient fine-tuning approach that allows models to progressively zoom in and localize small objects based on natural language expressions. 

\noindent \textbf{Future work and limitations.}
This work focused on localizing small objects in autonomous driving scenarios since detecting such objects is critical for ensuring safety and improves the overall reliability. Extending the proposed dataset to include more diverse environments and object types would remain an interesting future research direction.
In addition, extending this work to video data and applying PIZA tuning to architectures for video processing would also be a promising next step. 
We believe that this work contributed to the computer vision community from both dataset and technical perspectives.

\noindent {\small \textbf{Acknowledgment.}
This work was supported by JSPS KAKENHI 25K03135, 23H00490, JST CRONOS JPMJCS24K6, and JST BOOST Program JPMJBY24C5, and was carried out using the TSUBAME4.0 supercomputer at Institute of Science Tokyo.}\\
\vspace{-20pt}

{
    \small
    \bibliographystyle{ieeenat_fullname}
    \bibliography{main}
}
\def\result#1{\scalebox{0.93}[1]{#1}}
\def\resultz#1{\scalebox{0.98}[1]{#1}}
\def\tableMLLM{
\begin{table}[t]
\centering
\small
\vspace{0pt}
\setlength{\tabcolsep}{0.1pt}
\begin{tabular}{lcccc}
\toprule
Method & LLM & Val & Test-A & Test-B \\
\midrule
Zero-shot & \scalebox{0.93}[1]{Qwen2-VL-7B} &
\scalebox{0.97}[1]{0.2/0.8/0.0~~} &
\scalebox{0.97}[1]{0.3/1.1/0.1~~} &
\scalebox{0.97}[1]{0.1/0.3/0.0~~} \\
Full FT & \scalebox{0.93}[1]{Qwen2-VL-7B} &
\scalebox{0.97}[1]{1.9/6.4/0.6~~} &
\scalebox{0.97}[1]{2.4/7.9/0.8~~} &
\scalebox{0.97}[1]{1.2/4.2/0.4~~}
\\
LoRA  & \scalebox{0.93}[1]{Qwen2-VL-7B} &
\scalebox{0.97}[1]{3.8/12.4/1.4~~} &
\scalebox{0.97}[1]{5.0/15.5/2.0~~} &
\scalebox{0.97}[1]{2.6/8.7/0.9~}
\\
\scalebox{0.93}[1]{PIZA\hspace{-1pt}-LoRA} & \scalebox{0.93}[1]{Qwen2-VL-7B} & \scalebox{0.93}[1]{\underline{27.9}/\underline{49.4}/\underline{28.4}}& \scalebox{0.93}[1]{\underline{31.8}/\underline{52.4}/\underline{34.5}} & \scalebox{0.93}[1]{\underline{23.0}/\underline{42.6}/\underline{22.2}}\\
\midrule
Zero-shot & \scalebox{0.93}[1]{InternVL2.5-8B} &
\resultz{0.0/0.0/0.0} & 
\resultz{0.0/0.1/0.0} & 
\resultz{0.0/0.0/0.0}\\
Full FT & \scalebox{0.93}[1]{InternVL2.5-8B} &
\resultz{0.1/0.5/0.0} &
\resultz{0.1/0.5/0.0} &
\resultz{0.1/0.5/0.0} \\
LoRA & \scalebox{0.93}[1]{InternVL2.5-8B} &
\resultz{0.2/0.8/0.0} &
\resultz{0.2/0.8/0.1} &
\resultz{0.2/0.8/0.0} \\
\scalebox{0.93}[1]{PIZA\hspace{-1pt}-LoRA} & \scalebox{0.93}[1]{InternVL2.5-8B} & 
\result{\underline{20.7}/\underline{47.5}/\underline{19.2}~} & 
\result{\underline{25.4}/\underline{55.4}/\underline{24.5}~} &
\result{\underline{16.7}/\underline{40.2}/\underline{13.9}}
\\
\midrule
Zero-shot & \scalebox{0.88}[1]{LLaVA-NeXT-7B} &
\resultz{0.0/0.0/0.0} & 
\resultz{0.0/0.0/0.0} & 
\resultz{0.0/0.0/0.0}\\
Full FT & \scalebox{0.88}[1]{LLaVA-NeXT-7B} &
\resultz{0.1/0.3/0.0} & 
\resultz{0.1/0.2/0.0} & 
\resultz{0.1/0.2/0.0}\\
LoRA & \scalebox{0.88}[1]{LLaVA-NeXT-7B} &
\resultz{0.7/2.7/0.1} &  \resultz{0.7/2.5/0.1} & \resultz{0.6/2.2/0.2}\\
\scalebox{0.93}[1]{PIZA\hspace{-1pt}-LoRA} & \scalebox{0.88}[1]{LLaVA-NeXT-7B} &
\resultz{\underline{10.8}/\underline{27.7}/\underline{6.3}} & 
\resultz{\underline{12.0}/\underline{30.4}/\underline{7.2}} & 
\resultz{\underline{8.7}/\underline{23.1}/\underline{5.0}}\\
\bottomrule
\end{tabular}
\caption{Experiments with LLMs on SOREC (Train-L).
LoRA rank is set to 128. Each triplet of values indicates \scalebox{1.0}{mAcc}/\scalebox{1.0}{Acc$_{50}$}/\scalebox{1.0}{Acc$_{75}$}.}
\label{tab:mllm}
\vspace{-8pt}
\end{table}
}

\clearpage
\setcounter{page}{1}
\maketitlesupplementary

\begin{figure}[t]
\centering
\includegraphics[width=2.15\linewidth]{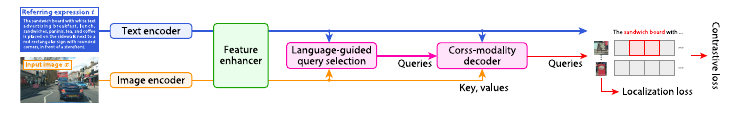}
\begin{minipage}{3\linewidth}
\captionsetup{margin=15em}
\caption{\label{fig:gdino_arc}\makebox[5.0\linewidth][l]{Architecture of GroundingDINO~\cite{Liu2024GroundingDINO, zhang2024mmgroundingdino}.}}
\end{minipage}
\vspace{0pt}
\end{figure}

\section*{A. Implementation details and analysis}
\label{sec:arc}
\noindent \textbf{Model architecture.}
Figure~\ref{fig:gdino_arc} shows the architecture of GroundingDINO~\cite{Liu2024GroundingDINO}, which we used as a backbone architecture in our experiments.
It consists of five components: a text encoder, an image encoder, a feature enhancer, a language guided query selection module, and a cross-modality decoder.
The BERT model is used as the text encoder.
The Swin transformer is used as the image encoder.
Figure~\ref{fig:fullft} shows the architecture of the feature enhancer and the decoder, to which we applied parameter-efficient fine-tuning methods.
Below, we describe details of each fine-tuning method.

\noindent \textbf{Full fine-tuning.}
Full fine-tuning uses all parameters as learnable parameters. The number of parameters for each component is listed in Table~\ref{tab:num_params}.

\begin{table}[t]
\centering
\vspace{120pt}
\includegraphics[width=\linewidth]{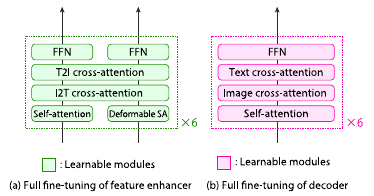}
\captionof{figure}{Block architectures of feature enhancer and decoder. Full fine-tuning updates all parameters.}
\label{fig:fullft}
\vspace{20pt}
\centering
\begin{tabular}{l|c|c}
\toprule
Module & Architecture & \#Params \\
\midrule
Text encoder & BERT & 108.9M\\
Image encoder & SwinT & 27.5M\\
Feature enhancer & Figure~\ref{fig:fullft}~\red{(a)} & 21.9M\\
Decoder & Figure~\ref{fig:fullft}~\red{(b)} & 11.1M\\
Others & - & 3.4M\\
\midrule
Total & - & 172.8M\\
\bottomrule
\end{tabular}
\captionof{table}{Number of parameters for each module.}
\label{tab:num_params}
\vspace{20pt}
\centering
\small
\setlength{\tabcolsep}{1pt}
\begin{tabular}{l|c|c|c|c}
\toprule
Method & \#Prm.  & Val & Test-A & Test-B \\
\midrule
PIZA-CoOp & \scalebox{0.97}[1]{0.9\MM} & \valmu{42.2}{39.1}{29.7}{26.3}\testABmu{43.0}{41.2}{34.2}{29.4}{36.7}{33.8}{24.3}{21.9} \\
\scalebox{0.95}[1]{w/o\hspace{1pt}emb.\hspace{1pt}insertion} & \scalebox{0.97}[1]{0.3\MM} & \valm{42.0}{38.7}{29.0}{26.1}\testABm{42.7}{40.9}{33.7}{29.3}{36.5}{33.2}{23.9}{21.6}\\
\scalebox{0.97}[1]{w/o\hspace{1pt}PIZA\hspace{1pt}module} & \scalebox{0.97}[1]{0.1\MM} & \valm{41.5}{36.1}{20.4}{20.2}\testABm{43.4}{40.1}{25.8}{24.2}{34.9}{29.6}{14.6}{15.5}\\
\midrule
$L=4$ & \scalebox{0.97}[1]{0.5\MM} & \valm{0.0}{\underline{39.2}}{29.3}{\underline{26.4}}\testABm{0.0}{41.0}{34.2}{29.5}{0.0}{\underline{33.9}}{\underline{24.4}}{\underline{21.9}}\\
$L=8$ & \scalebox{0.97}[1]{0.9\MM} & \valm{0.0}{39.1}{\underline{29.7}}{26.3}\testABm{0.0}{41.2}{34.2}{29.4}{0.0}{33.8}{24.3}{\underline{21.9}}\\
$L=16$ & \scalebox{0.97}[1]{1.7\MM} & \valm{0.0}{38.0}{29.1}{25.8}\testABm{0.0}{\underline{41.5}}{\underline{34.5}}{\underline{29.8}}{0.0}{32.5}{23.5}{21.1}\\
\bottomrule
\end{tabular}
\captionof{table}{
Ablation and hyperparameter studies for PIZA-CoOp. Train-S is used for training.
Each triplet of values indicates \scalebox{1.0}{mAcc}/\scalebox{1.0}{Acc$_{50}$}/\scalebox{1.0}{Acc$_{75}$}.
}
\label{tab:ablation_coop_appendix}
\vspace{-20pt}
\end{table}

\begin{figure*}[t]
\centering
\includegraphics[width=\linewidth]{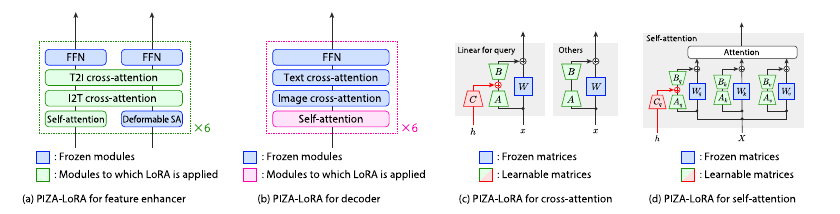}
\caption{PIZA-LoRA architecture.}
\label{fig:lora}
\centering
\includegraphics[width=\linewidth]{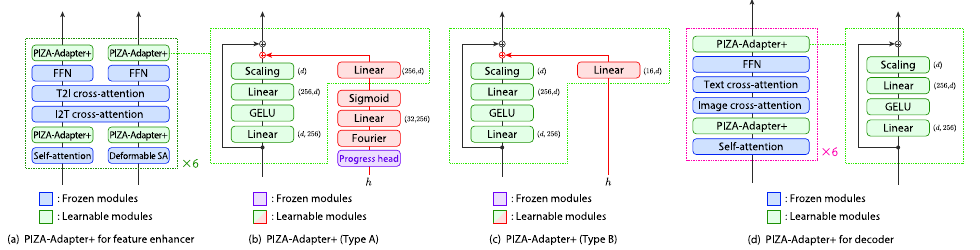}
\caption{PIZA-Adapter+ architecture. Types A and B are designed for training with a small dataset and a large dataset, respectively.}
\label{fig:adapter}
\end{figure*}

\noindent \textbf{PIZA-CoOp.}
CoOp is applied to the text encoder by prepending 16 learnable embeddings to input text prompt.
PIZA-CoOp further inserts zooming-step embeddings $\bm{h}$ between the prepended learnable embeddings and the text prompt via learnable linear layers $H$.
Specifically, $H$ consists of $L$ linear layers, $H_{1}, H_{2}, \cdots, H_{L}$, each of which is applied to $\bm{h}$ to obtain a sequence of embeddings whose length is $L = 8$.
During fine-tuning, all LayerNorm layers are also updated. 
Ablation and hyperparameter studies are shown in Table~\ref{tab:ablation_coop_appendix}.
Although we also tried larger values for $L$, they did not lead to improved performance on the validation and Test-B sets.
Overall, PIZA-CoOp did not surpass the results achieved by PIZA-LoRA and PIZA-Adapter+.

\noindent \textbf{PIZA-LoRA.}
We applied PIZA-LoRA to self-attention and cross-attention layers in the feature enhancer and decoder.
Figure~\ref{fig:lora} shows the detailed architecture.
For the feature enhancer, PIZA-LoRA is applied to its text-to-image cross-attention, image-to-text cross-attention, and self-attention modules.
For matrices to compute queries for the cross-attention modules, the zoom-step embedding is added to the LoRA bottleneck through a learnable matrix $C$.
The vanilla LoRA is applied to the other liner functions in this module because we observed that inserting zoom-step embedding did not improve the performance.
For self-attention modules, we applied PIZA-LoRA in the same way. During fine-tuning, all LayerNorm layers are also updated. Ablation and hyperparameter studies are shown in Table~\ref{tab:ablation_lora_appendix}. For PIZA-LoRA, increasing the rank to 64 slightly improved performance but did not achieve the performance level of PIZA-Adapter+ of Table~\ref{tab:main_results}.

\begin{table}[t]
\centering
\small
\setlength{\tabcolsep}{1pt}
\begin{tabular}{l|c|c|c|c}
\toprule
Method & \#Prm.  & Val & Test-A & Test-B \\
\midrule
PIZA-LoRA & \scalebox{0.97}[1]{1.5\MM} & \valmu{48.0}{44.7}{34.9}{30.9}\testABmu{48.9}{46.6}{39.2}{33.8}{42.0}{38.7}{28.9}{25.8}\\
\scalebox{0.95}[1]{w/o\hspace{1pt}emb.\hspace{1pt}insertion} & \scalebox{0.97}[1]{1.5\MM} & \valm{47.1}{43.9}{34.0}{30.2}\testABm{48.2}{46.4}{38.6}{33.5}{41.4}{38.1}{28.3}{25.3}\\
\scalebox{0.97}[1]{w/o\hspace{1pt}PIZA\hspace{1pt}module} & \scalebox{0.97}[1]{1.3\MM} & \valm{44.3}{38.5}{21.8}{21.6}\testABm{42.7}{43.1}{28.1}{26.2}{34.4}{32.5}{15.9}{17.0}\\
\midrule
$r = 64$ & \scalebox{0.97}[1]{5.1\MM} & \valm{48.2}{\underline{44.9}}{\underline{35.2}}{\underline{31.1}}\testABm{49.1}{\underline{47.3}}{\underline{40.0}}{\underline{34.4}}{41.6}{38.5}{28.9}{25.7}\\
$r = 16$ & \scalebox{0.97}[1]{1.5\MM} & \valm{48.0}{44.7}{34.9}{30.9}\testABm{48.9}{46.6}{39.2}{{33.8}}{42.0}{\underline{38.7}}{28.9}{\underline{25.8}}\\
$r = 4$ & \scalebox{0.97}[1]{0.6\MM} & \valm{48.0}{44.6}{34.7}{{30.8}}\testABm{48.2}{46.4}{39.1}{33.7}{41.7}{38.5}{\underline{29.0}}{\underline{25.8}} \\
\bottomrule
\end{tabular}
\caption{
Ablation and hyperparameter studies for PIZA-LoRA.
$r$ indicates the rank of the low-rank matrices. Train-S is used for training. Each triplet of values indicate \scalebox{1.0}{mAcc}/\scalebox{1.0}{Acc$_{50}$}/\scalebox{1.0}{Acc$_{75}$}.
}
\vspace{-8pt}
\label{tab:ablation_lora_appendix}
\end{table}

\begin{table*}[t]
\small
\centering
\setlength{\tabcolsep}{1pt}
\begin{tabular}{l|c|ccc|ccc|ccc|ccc|ccc|ccc}
\toprule
\multirow{3}{*}{Method} & \multirow{3}{*}{\#Prm.} & \multicolumn{9}{c|}{Train-S} & \multicolumn{9}{c}{Train-L}\\
& & \subsets\\
& & \metrics & \metrics & \metrics & \metrics & \metrics & \metrics\\
\midrule
PIZA-Adapter+\hspace{1pt}(Type A)& 3.5\MM & \val{56.8}{53.5}{41.8}{36.8}\testAB{61.6}{59.6}{50.1}{43.1}{49.5}{45.9}{34.1}{30.4} & \val{65.7}{59.2}{40.2}{37.0}\testAB{69.6}{64.7}{48.9}{43.1}{57.7}{50.3}{30.9}{29.7}\\
PIZA-Adapter+\hspace{1pt}(Type B)& 3.5\MM & \val{55.4}{51.6}{40.7}{35.6}\testAB{60.2}{58.0}{48.8}{41.8}{47.2}{43.6}{32.2}{28.6} & \val{66.5}{60.6}{42.9}{39.0}\testAB{69.9}{66.2}{51.7}{45.1}{58.6}{52.2}{33.6}{31.7}\\
\bottomrule
\end{tabular}
\caption{Comparison of PIZA-Adapter+ configurations.}
\label{tab:types}
\end{table*}

\noindent \textbf{PIZA-Adapter+.}
We applied PIZA-Adapter+ to the feature enhancer.
As shown in Figure~\ref{fig:adapter}, four PIZA-Adapter+ modules are inserted into each feature enhancer block in a post-adapter manner, {\it i.e.}, adapters are inserted after the feedforward networks and attention modules.
Each PIZA-Adapter+ module consists of either Type A in Figure~\ref{fig:adapter}~\red{(b)} or Type B in Figure~\ref{fig:adapter}~\red{(c)}.
Type A leverages the zoom progress value obtained from the frozen progress head of the PIZA module (the module to predict an EOS label with a progress value in Figure~\ref{fig:zoom_embedding}).
The time embedding module that originates from the stable diffusion~\cite{rombach2022sd}, consisting of a Fourier embedding and a small MLP, is then applied to the zoom progress value. Type B omits the progress head and uses the features extracted from the PIZA module. Comparison of Types A and B is shown in Table~\ref{tab:types}. Type A is particularly effective when a small size of dataset is used for training in experiments with Train-S.
Although Type B is seemingly more efficient than Type A, it requires more training data than Type A because it lacks the progress head that enlarges the variety of conditioning inputs.
We also designed PIZA-Adapter+ for the decoder in Figure~\ref{fig:adapter}~\red{(d)}, but this did not improve the performance.

\begin{table*}[t]
\small
\centering
\setlength{\tabcolsep}{1pt}
\begin{tabular}{l|c|ccc|ccc|ccc|ccc|ccc|ccc}
\toprule
\multirow{3}{*}{Method} & \multirow{3}{*}{\#Prm.} & \multicolumn{9}{c|}{Train-S} & \multicolumn{9}{c}{Train-L}\\
& & \subsets\\
& & \metrics & \metrics & \metrics & \metrics & \metrics & \metrics\\
\midrule
VPT & 0.1\MM & \val{0.0}{35.3}{19.9}{19.8}\testAB{0.0}{39.2}{25.5}{23.9}{0.0}{29.6}{14.3}{15.4} & \val{0.0}{39.8}{22.6}{22.5}\testAB{0.0}{44.5}{29.3}{27.2}{0.0}{33.5}{16.4}{17.4}\\
PIZA-VPT (Ours) & 0.6\MM & \val{0.0}{38.9}{29.6}{26.5}\testAB{0.0}{39.9}{33.1}{28.7}{0.0}{33.8}{24.9}{22.2} & \val{0.0}{43.3}{33.6}{29.7}\testAB{0.0}{46.7}{39.0}{33.7}{0.0}{37.9}{28.0}{25.0} \\
\bottomrule
\end{tabular}
\caption{Results for VPT and PIZA-VPT.}
\vspace{-8pt}
\label{tab:vpt}
\end{table*}

\noindent \textbf{PIZA-VPT.}
VPT~\cite{jia2022vpt} is applied to the image encoder by prepending learnable embeddings to input visual prompt as $G(\bm{x}, \bm{t}) = F([\bm{e}, \bm{x}], \bm{t})$, where $\bm{e}$ is a sequence of learnable embeddings of length 16.
PIZA-VPT inserts the zooming-step embeddings $\bm{h}$ as $G_{\piza}(\bm{x}, \bm{t}) = F([\bm{e}, H(\bm{h}), \bm{x}], \bm{t})$, where $H$ is the module consisting of $L$ linear layers, similar to PIZA-CoOp.
As shown in Table~\ref{tab:vpt}, PIZA-VPT outperformed VPT.
Table~\ref{tab:ablation_vpt_appendix} shows ablation and hyperparameter studies. Similar to CoOp, PIZA-VPT did not surpass the results achieved by PIZA-LoRA and PIZA-Adapter+.

\noindent \textbf{Experiments in LMMs}
We conducted experiments with three large multimodal models (LMMs) that can perform the REC task: Qwen2-VL-7B, InternVL-2.5-8B ($n_{\text{max}}\hspace{-3pt}=\hspace{-3pt}24$ for dynamic resolution), and LLaVA-NeXT-Mistral-7B. We utilized LoRA tuning with their prompts for REC.
As shown in Table \ref{tab:mllm}, our PIZA approach unlocks their abilities to perform REC on small objects and significantly boosts the performance.
While full fine-tuning and LoRA tuning led to slight improvements, their performance was significantly lower compared to PIZA-LoRA.

\begin{table}[t]
\centering
\small
\setlength{\tabcolsep}{1pt}
\begin{tabular}{l|c|c|c|c}
\toprule
Method & \#Prm.  & Val & Test-A & Test-B \\
\midrule
PIZA-VPT & \scalebox{0.97}[1]{0.6\MM} & \valmu{0.0}{38.9}{29.6}{26.5}\testABmu{0.0}{39.9}{33.1}{28.7}{0.0}{33.8}{24.9}{22.2} \\
\scalebox{0.95}[1]{w/o\hspace{1pt}emb.\hspace{1pt}insertion} & \scalebox{0.97}[1]{0.4\MM} & \valm{0.0}{37.9}{29.4}{25.9}\testABm{0.0}{39.0}{32.7}{28.2}{0.0}{32.8}{24.3}{21.8}\\
\scalebox{0.97}[1]{w/o\hspace{1pt}PIZA\hspace{1pt}module} & \scalebox{0.97}[1]{0.1\MM} & \valm{0.0}{35.3}{19.9}{19.8}\testABm{0.0}{39.2}{25.5}{23.9}{0.0}{29.6}{14.3}{15.4}\\
\midrule
$L=4$ & \scalebox{0.97}[1]{0.5\MM} & \valm{0.0}{37.3}{28.9}{25.7}\testABm{0.0}{38.7}{32.6}{27.9}{0.0}{32.7}{24.2}{21.7}\\
$L=8$ & \scalebox{0.97}[1]{0.6\MM} & \valmu{0.0}{38.9}{29.6}{26.5}\testABmu{0.0}{39.9}{33.1}{28.7}{0.0}{33.8}{24.9}{22.2}\\
$L=16$ & \scalebox{0.97}[1]{0.8\MM} & \valm{0.0}{38.0}{\underline{29.6}}{26.0}\testABm{0.0}{38.5}{32.1}{27.8}{0.0}{32.8}{24.3}{21.7}\\
\bottomrule
\end{tabular}
\captionof{table}{
Ablation and hyperparameter studies for PIZA-VPT.
Train-S is used for training.
Each triplet of values indicates \scalebox{1.0}{mAcc}/\scalebox{1.0}{Acc$_{50}$}/\scalebox{1.0}{Acc$_{75}$}.
}
\label{tab:ablation_vpt_appendix}
\vspace{-8pt}
\end{table}

\begin{table}[t]
\centering
\small
\setlength{\tabcolsep}{1pt}
\begin{tabular}{l|c|ccc}
\toprule
Method & \#Prm. & Val & TestA & TestB\\
\midrule
Zero-shot & 0 & 50.4 & 57.2 & 43.2 \\
Full fine-tuning & 173\MM & 89.2 & 91.9 & 86.0 \\
\midrule
Adapter+ & 3.5\MM & 86.9 & 89.6 & 83.3\\
PIZA-Adapter+\hspace{1pt}(Ours) & 3.5\MM & 87.4 & 90.2 & 84.0\\
\bottomrule
\end{tabular}
\caption{Results on RefCOCO. Each value indicates Acc$_{50}$.}
\label{tab:refcoco_appendix}
\vspace{-13pt}
\end{table}

\noindent \textbf{RefCOCO.}
We also ran experiments on RefCOCO to demonstrate that methodological improvements with PIZA modules for small objects doesn't significantly impact the performance for objects of other sizes.
RefCOCO is one of the first dataset for referring expression comprehension~\cite{yu2016refcoco,mao2016refcocog,kazemzadeh2014referitgame, plummer2015flickr30k}. Unlike the SOREC dataset for small objects, RefCOCO concentrates on referring expressions for objects that occupy a relatively large portion of the images in MS-COCO~\cite{lin2014coco}. RefCOCO became a commonly-used benchmark of the referring expression comprehension task for a long time.
As shown in Table~\ref{tab:refcoco_appendix}, our PIZA-Adapter+ doesn't greatly decrease the accuracy on RefCOCO. PIZA-Adapter+ outperforms Adapter+ because the small learnable PIZA module helps improve the performance in RefCOCO. For larger objects, the \texttt{[EOS]} token was predicted after the first inference step.

\tableMLLM

\begin{figure*}[t]
    \centering
    \includegraphics[width=\linewidth]{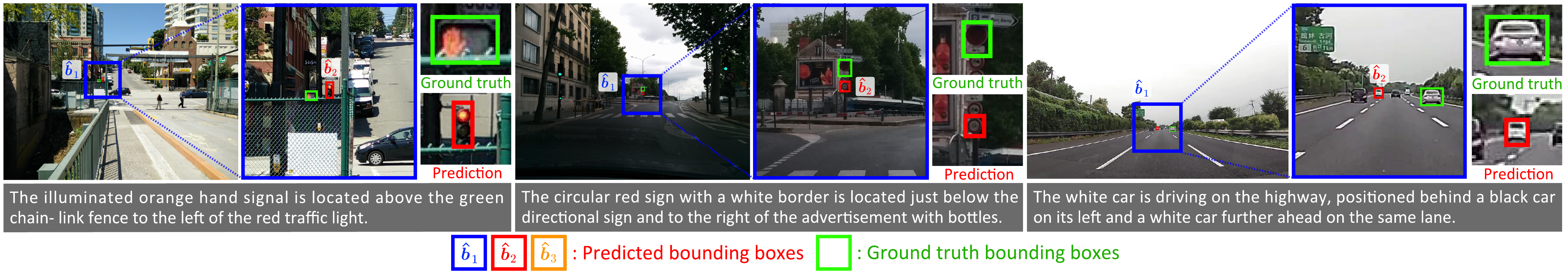}
    \caption{Failure cases.}
    \label{fig:error_analysis}
\end{figure*}

\def\qcaption{Qualitative examples.}

\begin{figure*}[t]
    \centering
    \includegraphics[width=\linewidth]{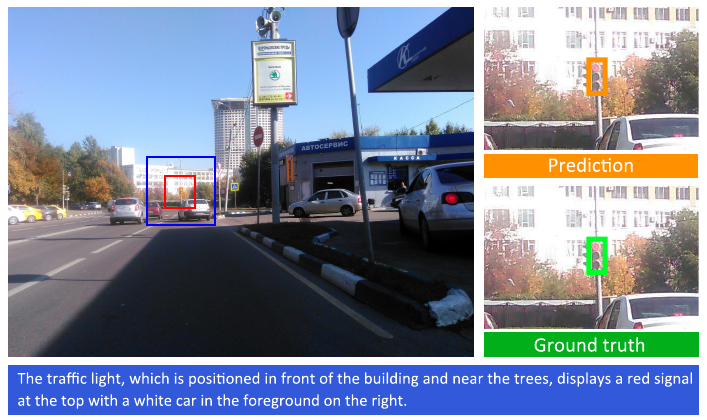}
    \caption{\qcaption}
    \label{fig:qualitative0}
\end{figure*}

\noindent \textbf{Extended training dataset.}
The extended training dataset $\mathcal{E}$ consists of ground truth search processes  $P^{*}$. The average length of $P^{*}$ was 2.11, indicating that, in most cases, two zooming steps are sufficient to localize objects in the SOREC dataset via fine-tuning with PIZA.
The hyperparameters $\lambda_{1}, \lambda_{2}$ of the weighting function were optimized for each target bounding box to ensure that the minimum size of the input image is larger than 450 pixels, as we empirically found that including smaller cropped images degrades the performance. Specifically, the parameters are first set to $\lambda_{1} = 1.0$ and $\lambda_{2} = 1.0$, and then if the edge length of the second-to-last bounding box is smaller than 450 pixels, we resample the search process by multiplying $\lambda_{2}$ by 1.1, iteratively until the edge length exceeds the threshold.

\noindent \textbf{Error analysis.}
We analyzed failure cases, as shown in Figure~\ref{fig:error_analysis}.
The results indicate that localizing objects occluded by other objects or placed in close proximity to similar objects remains challenging.
Creating datasets with 8K or higher resolution images, which may require additional steps, is also left for future work.

\begin{figure*}[t]
    \centering
    \includegraphics[width=\linewidth]{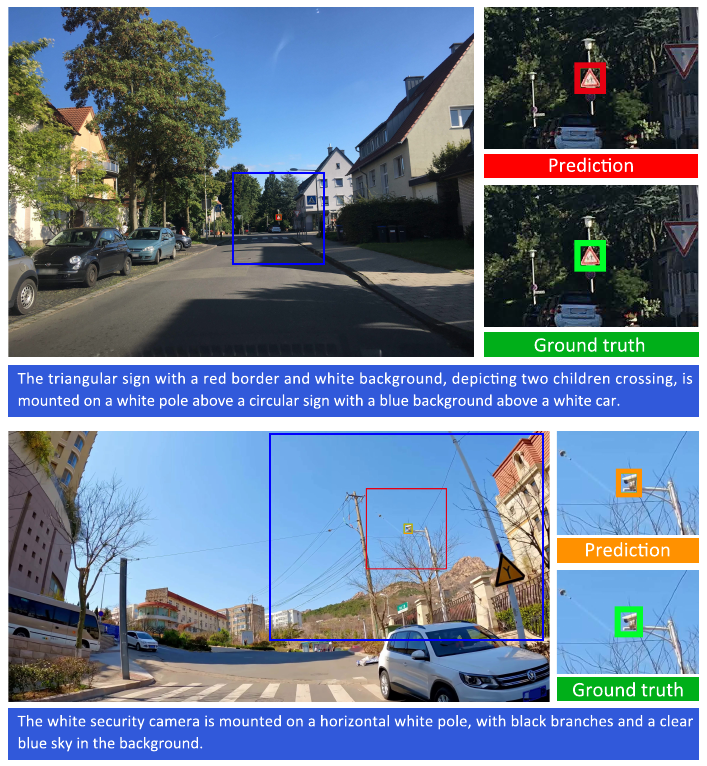}
    \caption{\qcaption}
    \label{fig:qualitative1}
\end{figure*}

\begin{figure*}[t]
    \centering
    \includegraphics[width=\linewidth]{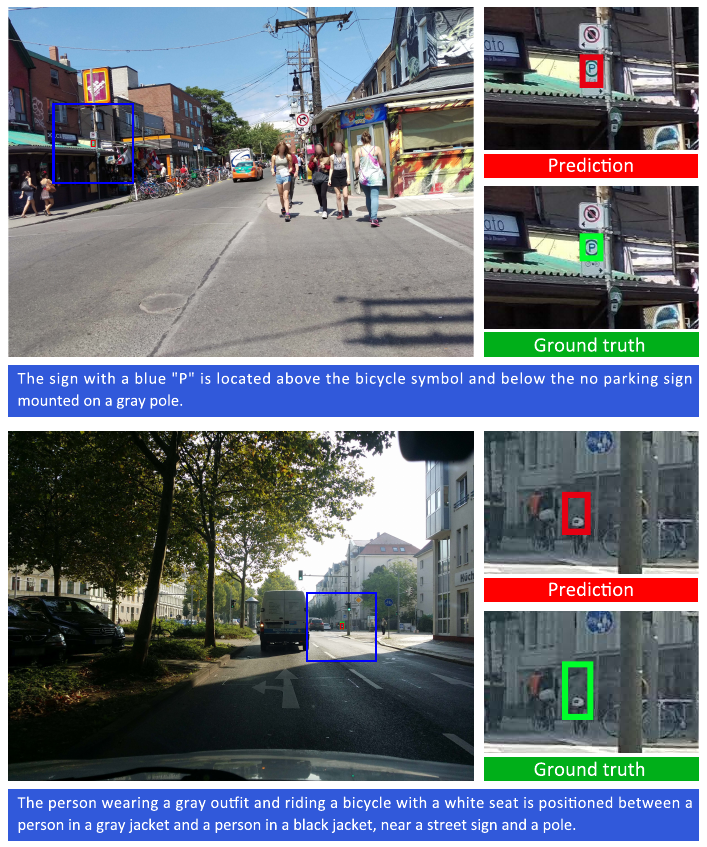}
    \caption{\qcaption}
    \label{fig:qualitative2}
\end{figure*}

\begin{figure*}[t]
    \centering
    \includegraphics[width=\linewidth]{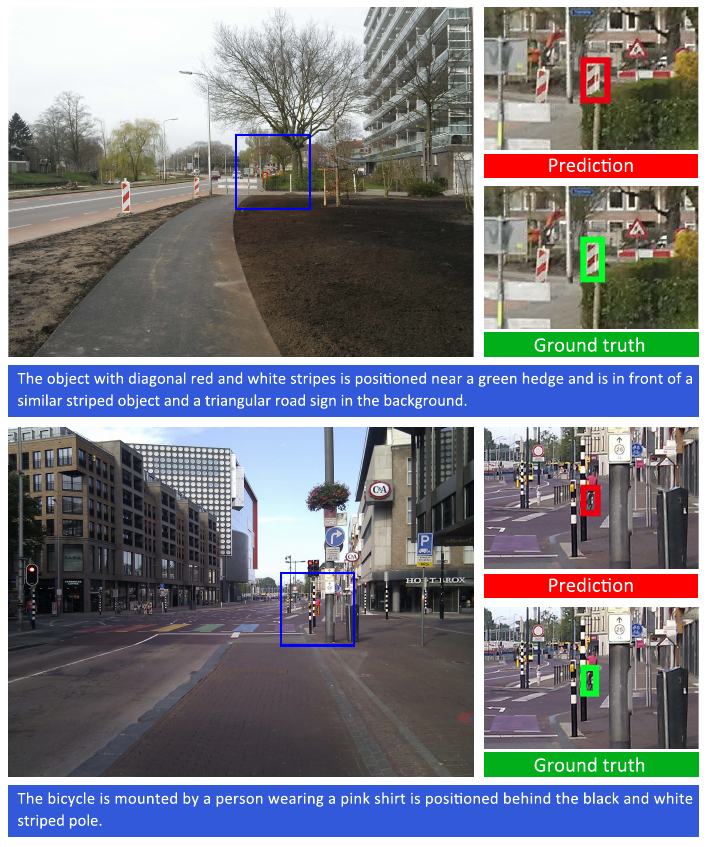}
    \caption{\qcaption}
    \label{fig:qualitative3}
\end{figure*}

\begin{figure*}[t]
    \centering
    \includegraphics[width=\linewidth]{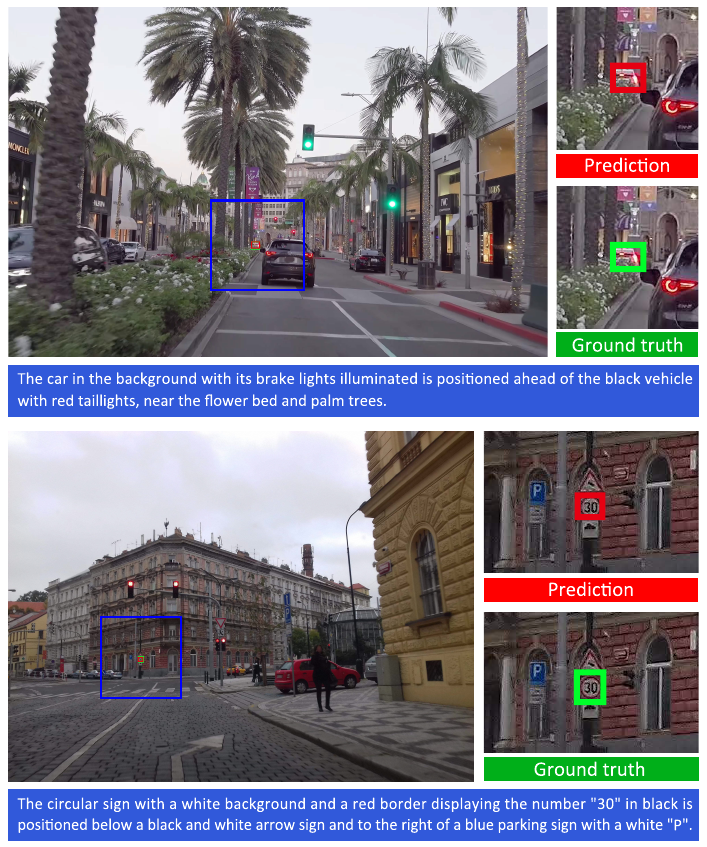}
    \caption{\qcaption}
    \label{fig:qualitative4}
\end{figure*}

\begin{figure*}[t]
    \centering
    \includegraphics[width=\linewidth]{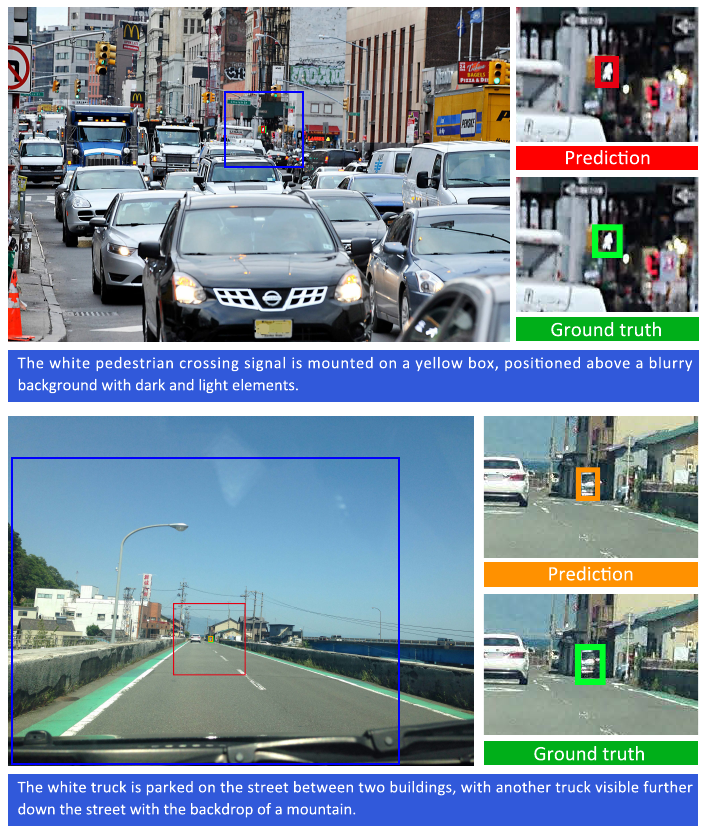}
    \caption{\qcaption}
    \label{fig:qualitative6}
\end{figure*}

\noindent \textbf{Qualitative examples.}
Figures~\ref{fig:qualitative0} to \ref{fig:qualitative6} show qualitative examples.
The predicted bounding boxes, $\bm{\hat{b}}_{1}, \bm{\hat{b}}_{2},$ and $ \bm{\hat{b}}_{3}$, are colored in blue, red, and orange, respectively, in each figure. The final predictions, $\bm{\hat{b}}_{\hat{T}}$, where $\hat{T} = 2$ or $3$, are compared with the ground truth bounding boxes in green. As shown, our method successfully localizes extremely small target objects.

\end{document}